%% file: main.tex
\documentclass[11pt, a4paper, logo, copyright]{googlecloud}

\pdftrailerid{redacted}

\makeatletter
\renewcommand\bibentry[1]{\nocite{#1}{\frenchspacing\@nameuse{BR@r@#1\@extra@b@citeb}}}
\makeatother

\usepackage{kantlipsum, lipsum}
\usepackage{dsfont}

\usepackage[authoryear, sort&compress, round]{natbib}

\usepackage[utf8]{inputenc} %
\usepackage[T1]{fontenc}    %
\usepackage{url}            %
\usepackage{booktabs}       %
\usepackage{nicefrac}       %
\usepackage{microtype}      %
\usepackage{amsmath}
\usepackage{graphicx}
\usepackage{multicol}
\usepackage{hyperref}       %
\usepackage[nameinlink]{cleveref}
\usepackage{bbm}
\usepackage{multirow}
\usepackage{soul}
\usepackage{floatrow}
\usepackage{float}
\usepackage{wrapfig}
\usepackage{blindtext}
\usepackage{tablefootnote}
\usepackage{amsfonts}
\usepackage[flushleft]{threeparttable}
\usepackage{bbding}
\usepackage{xcolor}
\usepackage{xspace}
\usepackage{bm}
\usepackage{arydshln}
\usepackage{enumitem}
\usepackage{setspace}
\usepackage{color}
\usepackage{algorithm}
\usepackage{longtable}

\usepackage[normalem]{ulem}
\usepackage{ulem}
\usepackage[nomargin,inline,marginclue,draft]{fixme}
\usepackage{balance}
\usepackage{verbatim}
\usepackage{diagbox}
\usepackage{changepage}
\usepackage{amssymb}
\usepackage{pifont}
\usepackage{array}   %

\usepackage{hyperref}
\usepackage{url}
\usepackage{amsmath} 
\usepackage{enumitem}
\usepackage{booktabs}
\usepackage{graphicx}
\usepackage{adjustbox}
\usepackage{multirow}
\usepackage{algorithm}
\usepackage{algpseudocode}
\usepackage{subcaption}
\definecolor{nb}{HTML}{006EB8}
\definecolor{red_}{HTML}{cd6155}
\definecolor{green_}{HTML}{52be80}
\usepackage{hyperref}
\hypersetup{
  colorlinks   = true,    %
  urlcolor     = nb,    %
  linkcolor    = nb,    %
  citecolor    = nb      %
}

\usepackage{booktabs,makecell}

\usepackage{tcolorbox}

\newtcolorbox{prompt}[2][]{
    colback=white,
    colframe=gray!45,
    fonttitle=\bfseries,
    coltitle=black,
    sharp corners,
    title=#2,
    #1
}

\tcbset{
    promptstyle/.style={
        enhanced,
        colback=white,
        colframe=black,
        colbacktitle=gray!20,
        width=0.98\linewidth, 
        coltitle=black,
        rounded corners,
        sharp corners=north,
        boxrule=0.5pt,
        drop shadow=black!50!white,
        attach boxed title to top left={
            xshift=-2mm,
            yshift=-2mm
        },
        title code={%
            \begin{minipage}{0.8\linewidth}
                \centering\thetitle
            \end{minipage}
        },
        boxed title style={
            rounded corners,
            size=small,
            colback=gray!20,
        },
        fonttitle=\normalsize\bfseries,
        before upper={\parindent15pt}
    }
}

\newtcolorbox{promptbox}[1][]{
    promptstyle,
    title=Prompt,
    #1
}

\usepackage{tikz,pgfplots}
\pgfplotsset{compat=1.18}
\usepackage{xcolor}
\usetikzlibrary{arrows.meta,calc,positioning,fit,shadows.blur,decorations.pathreplacing}

\definecolor{pHeaderL}{RGB}{238,242,255}
\definecolor{pHeaderR}{RGB}{253,242,248}
\definecolor{pOursBg}{RGB}{224,247,241}
\definecolor{pOursStroke}{RGB}{134,239,172}
\definecolor{pCell}{RGB}{248,250,252}
\definecolor{pStroke}{RGB}{229,231,235}
\definecolor{pBadgeOK}{RGB}{220,252,231}
\definecolor{pBadgeNG}{RGB}{254,226,226}
\definecolor{pText}{RGB}{31,41,55}
\definecolor{pSub}{RGB}{75,85,99}
\definecolor{pWarn}{RGB}{239,68,68}
\definecolor{pOK}{RGB}{16,185,129}
\definecolor{pGold}{RGB}{251,191,36}

\definecolor{StarCoral}{HTML}{F58A7A} 
\definecolor{StarTeal}{HTML}{4FB5A5} 
\definecolor{StarIndigo}{HTML}{4C6EDB}  
\definecolor{StarYellow}{HTML}{EBC065}

\usepackage{tabularx}

\usepackage{amsmath}
\usepackage{wasysym}
\usepackage{times}
\usepackage{latexsym}
\usepackage{tcolorbox}           %
\tcbuselibrary{skins,breakable,listings,theorems}
\usepackage[T1]{fontenc}
\usepackage{inconsolata}
\usepackage{graphicx}
\usepackage{microtype}
\usepackage{amssymb}
\usepackage{mathtools}
\usepackage{amsthm}
\usepackage{multirow}
\usepackage{makecell}
\usepackage{booktabs}
\usepackage{array}
\usepackage{longtable}
\usepackage{fontawesome}
\usepackage{centernot}

\usepackage{xcolor}

\usepackage{listings}

\tcbset{
    promptstyle/.style={
        enhanced,
        colback=white,
        colframe=black,
        colbacktitle=gray!20,
        width=0.98\linewidth, 
        coltitle=black,
        rounded corners,
        sharp corners=north,
        boxrule=0.5pt,
        drop shadow=black!50!white,
        attach boxed title to top left={
            xshift=-2mm,
            yshift=-2mm
        },
        title code={%
            \begin{minipage}{0.8\linewidth}
                \centering\thetitle
            \end{minipage}
        },
        boxed title style={
            rounded corners,
            size=small,
            colback=gray!20,
        },
        fonttitle=\normalsize\bfseries,
        before upper={\parindent15pt}
    }
}

\definecolor{codegreen}{rgb}{0,0.6,0}
\definecolor{codegray}{rgb}{0.5,0.5,0.5}
\definecolor{codepurple}{rgb}{0.58,0,0.82}
\definecolor{backcolour}{rgb}{0.95,0.95,0.92}

\lstdefinestyle{mystyle}{
    backgroundcolor=\color{white},
    commentstyle=\color{codegreen},
    keywordstyle=\color{magenta},
    numberstyle=\tiny\color{codegray},
    stringstyle=\color{codepurple},
    basicstyle=\sffamily\footnotesize,
    breakatwhitespace=false,         
    breaklines=true,                 
    captionpos=b,                    
    keepspaces=true,                 
    numbers=none,
    numbersep=5pt,                  
    showspaces=false,                
    showstringspaces=false,
    showtabs=false,                  
    tabsize=2
}
\usepackage{siunitx}
\usepackage{xcolor}

\usepackage{minted} 
\setminted{tabsize=2, breaklines=true, breakanywhere=true}
\usepackage{algpseudocode}

\usepackage{tcolorbox}

\title{CAST: Modeling Visual State Transitions for Consistent Video Retrieval}

\correspondingauthor{yliu858@ucsc.edu, yanjiao@google.com}

\renewcommand{\today}{}

\author[1 2 *]{Yanqing Liu}
\author[1 3]{Yingcheng Liu}
\author[1]{Fanghong Dong}
\author[1]{Budianto Budianto}
\author[2]{Cihang Xie}
\author[1]{Yan Jiao}

\affil[1]{Google}
\affil[2]{University of California, Santa Cruz}
\affil[3]{MIT}

\begin{abstract}
As video content creation shifts toward long-form narratives, composing short clips into coherent storylines becomes increasingly important. However, prevailing retrieval formulations remain context-agnostic at inference time, prioritizing local semantic alignment while neglecting the state and identity consistency. To address this structural limitation, we formalize the task of \textbf{Consistent Video Retrieval (CVR)} and introduce a diagnostic benchmark spanning YouCook2, COIN, and CrossTask.
We propose \textbf{CAST (Context-Aware State Transition)}, a lightweight, plug-and-play adapter compatible with diverse frozen vision-language embedding spaces. By predicting a state-conditioned residual update ($\Delta$) from visual history, CAST introduces an explicit inductive bias for latent state evolution.
Extensive experiments show that CAST improves performance on YouCook2 and CrossTask, remains competitive on COIN, and consistently outperforms zero-shot baselines across diverse foundation backbones. Furthermore, CAST provides a useful reranking signal for black-box video generation candidates (e.g., from Veo), promoting more temporally coherent continuations.
\end{abstract}

\begin{document}

\maketitle

\input{sec/1_intro}

\input{sec/2_method}

\input{sec/3_exp}
\input{sec/4_related_work}
\input{sec/5_limit}
\input{sec/6_conclusion}

\bibliographystyle{abbrvnat}
\nobibliography*
\bibliography{ref}

\clearpage

\appendix
\input{sec/appendix}

\end{document}

%% file: sec/1_intro.tex
\section{Introduction}
Video content creation is increasingly shifting from isolated clips to structured multi-step compositions, driven by recent advances in large-scale video generation models~\cite{villegas2022phenaki, openai2024sora, team2025kling, veo2024}. A central challenge in this setting is \emph{composition}: selecting and ordering short video segments to form a temporally coherent sequence. Procedural activities, such as cooking tutorials (e.g., ``wash'' $\rightarrow$ ``cut'' $\rightarrow$ ``cook''), provide a concrete and widely studied instance of this problem, where each step requires retrieving the appropriate next segment from a candidate set. Yet existing retrieval approaches treat these selections independently, prioritizing local semantic alignment while often producing fragmented narratives.

\begin{figure*}[t]
    \centering
    \includegraphics[width=\linewidth]{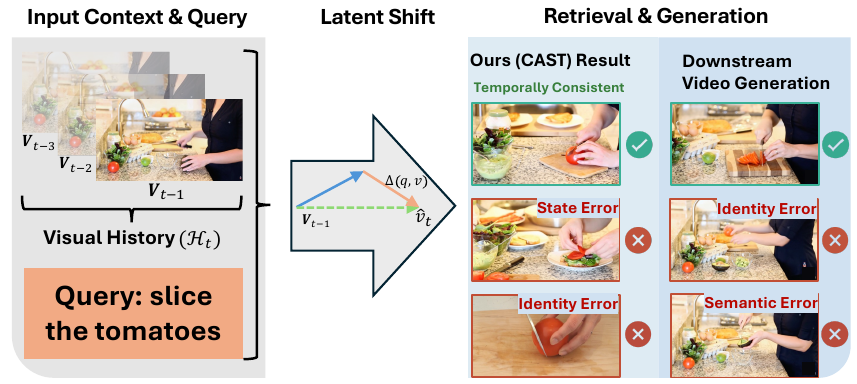}
    
    \caption{Given a context clip and instruction, standard retrieval often returns semantically relevant but temporally incoherent clips, yielding \textbf{State Errors} or \textbf{Identity Errors}. In contrast, \textbf{CAST} models the state transition ($\Delta$) to retrieve a causally plausible continuation and rerank generation candidates toward more coherent continuations.}
    \label{fig:teaser}
\end{figure*}

This limitation is rooted in the design of standard video retrieval itself. To enable scalable indexing, prevailing formulations encode clips independently and therefore operate under a context-agnostic inference paradigm~\cite{miech2019howto100m, luo2021clip4clip, bain2021frozen, zhao2024videoprism, wang2022internvideo}. In practice, this design induces two recurrent forms of incoherence. The first is identity inconsistency: abrupt shifts in actor, environment, or visual style across consecutive clips. The second, particularly salient in procedural settings, is state inconsistency: violations of procedural causality. For example, as illustrated in Figure~\ref{fig:teaser}, given a context clip of a person preparing ingredients, a query for ``slice the tomatoes'' may retrieve a later step showing \textit{already sliced} tomatoes being plated (State Error), or a clip of a different person cutting on a mismatched \textit{wooden board} (Identity Error), rather than the correct continuation of the original actor. Thus, although current methods achieve strong semantic recall on standard video retrieval benchmarks such as MSR-VTT~\cite{xu2016msr}, these metrics often fail to capture the temporal consistency required for coherent storytelling. Crucially, this failure mode does not primarily arise from weak visual or textual representations; rather, it reflects the absence of an explicit inductive bias for modeling state evolution under contextual constraints.

To address this gap, we introduce \textbf{Consistent Video Retrieval (CVR)}. Moving beyond standard text-to-video retrieval, CVR recasts retrieval as a \textit{context-aware} inference problem: given a sequence of preceding clips and a text instruction, the model must retrieve a target clip that is not only semantically aligned with the instruction, but also consistent with the preceding visual state and identity cues. To evaluate this setting rigorously, we establish a protocol derived from procedural datasets: YouCook2~\cite{zhou2018towards}, COIN~\cite{tang2019coin}, and CrossTask~\cite{zhukov2019cross}. While our broader motivation is long-horizon visual coherence, we focus on procedural activities as a controlled testbed. Unlike open-ended movies, procedural activities exhibit strict temporal ordering constraints, where steps cannot be arbitrarily permuted~\cite{alayrac2016unsupervised, zhou2018towards, zhukov2019cross}. These constraints often manifest as persistent changes in the underlying visual state (e.g., an egg cannot be un-cracked). Accordingly, unlike standard benchmarks dominated by semantically distinct distractors, our protocol constructs hard negatives that match the instruction while differing in state or identity, making retrieval metrics explicitly sensitive to consistency errors.

Building on this formulation, we propose \textbf{CAST} (\textbf{C}ontext-\textbf{A}ware \textbf{S}tate \textbf{T}ransition), motivated by the observation that actions in procedural narratives are better understood as latent visual state transitions than as static semantic labels. CAST therefore models procedural progression as a sequence of state-conditioned transitions. In contrast to context-agnostic matching, CAST uses the text instruction to predict a \textbf{residual vector ($\Delta$)} in the embedding space. This residual updates the current state embedding, modifying procedural attributes while retaining identity-relevant information from the anchor state through the additive connection~\cite{he2016deep}. Formally, this yields the retrieval target $\hat{v}_t \approx v_{t-1} + \Delta(v_{t-1}, q_t)$ (see \S\ref{subsec:cast_model} for full formulation), enabling the model to search for a causally plausible continuation rather than merely a semantic match. CAST is implemented as a lightweight trainable adapter~\cite{houlsby2019parameter, hu2022lora, gao2023llama} atop frozen foundation models, while remaining plug-and-play at inference time. Extensive experiments on our CVR benchmark show that this transition-based formulation consistently outperforms context-agnostic baselines, yielding stronger state discrimination while retaining identity cues and semantic recall. Beyond retrieval, we further provide preliminary evidence that CAST can serve as a useful reranking signal for selecting more coherent candidates from black-box video generation models (e.g., Veo).

In summary, our contributions are threefold:
\begin{itemize}
    \item We formalize Consistent Video Retrieval (CVR) and introduce a diagnostic benchmark that isolates state and identity consistency through hard negatives.
    \item We propose CAST, a lightweight trainable adapter that predicts latent state transitions $\Delta$ to support causally plausible retrieval under procedural context.
    \item Beyond retrieval, we provide preliminary evidence that CAST can serve as a useful reranking signal for black-box video generation candidates (e.g., from Veo).
\end{itemize}

%% file: sec/2_method.tex
\begin{figure*}[t]
    \centering
    \includegraphics[width=0.95\linewidth]{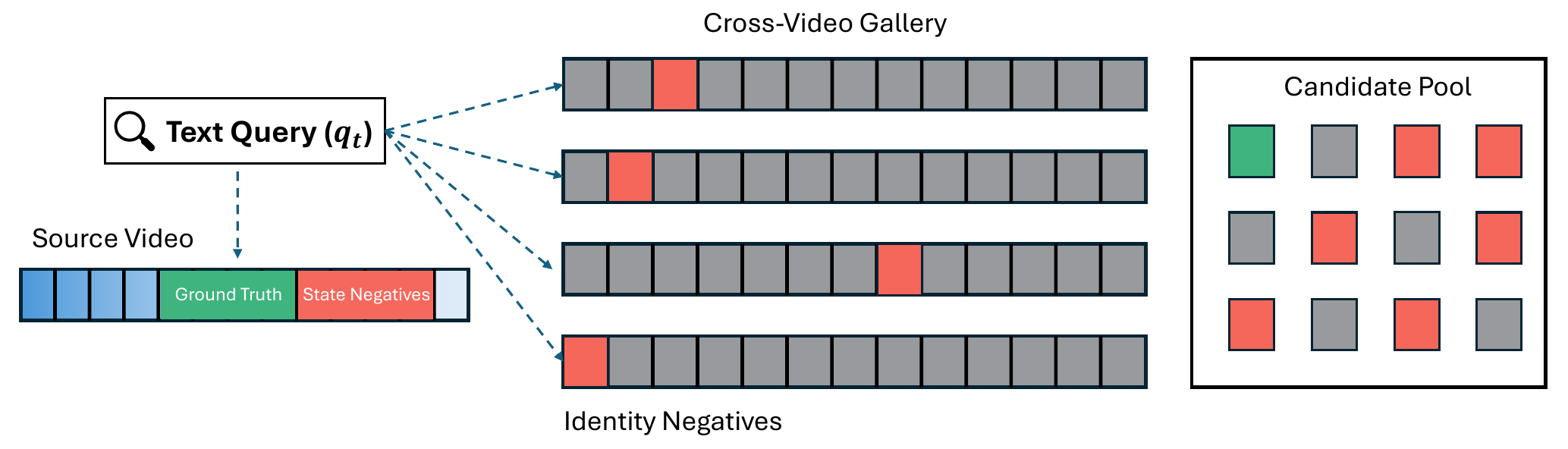}
    
    \caption{\textbf{Illustration of our CVR benchmark protocol.} In contrast to standard global retrieval, our benchmark introduces \textbf{State Negatives} (temporally misaligned clips from the same video) and \textbf{Identity Negatives} (appearance-misaligned clips from different videos) to diagnose consistency failures beyond semantic matching.}
    \label{fig:benchmark}
    
    \vspace{-10pt} 
\end{figure*}
\section{Method}
\label{sec:method}
\begin{figure*}[t] 
    \centering
    \includegraphics[width=0.45\linewidth]{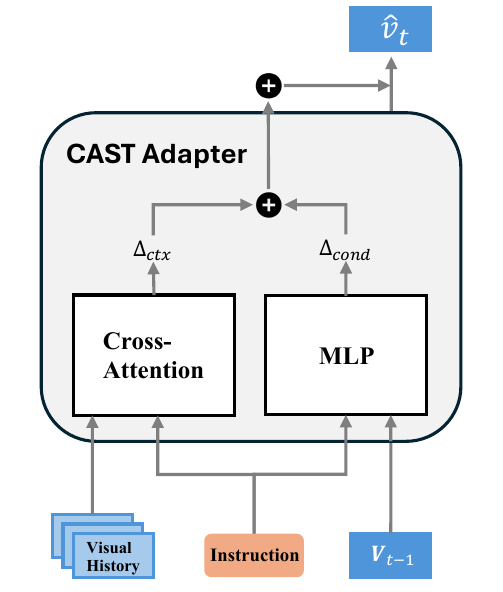}
    
    \caption{\textbf{Overview of the CAST adapter.} CAST operates as a lightweight adapter that aggregates visual history $\mathcal{H}_t$, anchor state $v_{t-1}$, and instruction $q_t$. Through a dual-path transition predictor, it estimates a residual update $\Delta$ that encourages causally consistent state evolution while retaining identity cues through the residual connection.}
    \label{fig:architecture}
    
\end{figure*}

We first introduce our diagnostic benchmark protocol (Figure~\ref{fig:benchmark}; \S\ref{subsec:eval_protocol}) for evaluating state and identity consistency in retrieval. We then present CAST (Figure~\ref{fig:architecture}; \S\ref{subsec:cast_model}), a lightweight state-transition adapter that addresses these failure modes via residual transition modeling.

\subsection{Problem Formulation}
\label{subsec:task_definition}

\noindent\textbf{Standard Text-to-Video Retrieval.}
Given a video gallery $\mathcal{V} = \{v_i\}_{i=1}^N$ and a text query $q$, standard retrieval seeks the clip $v^* \in \mathcal{V}$ with the highest semantic similarity to $q$. Formally, this amounts to maximizing similarity in the joint embedding space:
\begin{equation}
    v^* = \operatorname*{argmax}_{v \in \mathcal{V}} \, \text{sim}(f_t(q), f_v(v)),
    \label{eq:standard_retrieval}
\end{equation}
where $f_t$ and $f_v$ are pre-trained encoders (e.g., CLIP), and $\text{sim}(\cdot)$ denotes a similarity metric such as cosine similarity. Equivalently, this formulation models $P(v \mid q)$. Crucially, each query is treated independently. In procedural narratives, however, this independence assumption is fundamentally ill-suited: it ignores the preceding visual context and therefore may retrieve clips that are semantically relevant to the query yet inconsistent with the evolving event state.

\noindent\textbf{Consistent Video Retrieval.}
To overcome the structural limitation of standard retrieval, we reformulate the task as a sequential, context-aware inference problem in which the target clip $v_t$ is conditioned on both the current instruction $q_t$ and the visual history. Let $\mathcal{H}_t = \{v_{t-L}, \dots, v_{t-1}\}$ denote the recent narrative history preceding step $t$, where $L$ is the context window length. Our objective is to maximize the conditional probability:
\begin{equation}
    v_t^* = \operatorname*{argmax}_{v \in \mathcal{V}} \, P(v \mid \mathcal{H}_t, q_t).
    \label{eq:cvr_retrieval}
\end{equation}
Eq.~\ref{eq:cvr_retrieval} imposes two constraints: $v_t^*$ must remain locally aligned with the instruction $q_t$ while also preserving state and identity consistency with $\mathcal{H}_t$. In practice, we instantiate Eq.~\ref{eq:cvr_retrieval} by predicting a target embedding $\hat{v}_t$ and comparing it against candidates in the shared embedding space. This shift from context-agnostic matching in Eq.~\ref{eq:standard_retrieval} to context-aware inference in Eq.~\ref{eq:cvr_retrieval} motivates CAST.

\subsection{Benchmark Construction}
\label{subsec:eval_protocol}

\noindent\textbf{Limitations of Existing Benchmarks.}
Standard video retrieval benchmarks, such as MSR-VTT~\cite{xu2016msr}, MSVD~\cite{chen2011collecting}, and DiDeMo~\cite{anne2017localizing}, typically adopt a global retrieval protocol that ranks each query against the full video gallery. Because most gallery clips differ substantially in semantic content, models can perform well by relying on coarse-grained visual cues, such as objects or scenes, without modeling temporal consistency. As a result, these benchmarks are insufficiently diagnostic: they do not explicitly penalize \textit{state} or \textit{identity} inconsistencies.

\noindent\textbf{Procedural Data and Context.}
To address this gap, we construct the CVR benchmark using three procedural datasets: YouCook2~\cite{zhou2018towards}, COIN~\cite{tang2019coin}, and CrossTask~\cite{zhukov2019cross}. These datasets exhibit strong causal dependencies ($step_{t-1} \rightarrow step_t$), making them well-suited for evaluating temporal consistency. CrossTask further broadens the evaluation scope beyond cooking. We construct samples using a sliding-window strategy. Annotated step segments serve as the basic clip units, and the corresponding step text is treated as $q_t$. For each query step $q_t$, the context $\mathcal{H}_t$ consists of a variable-length sequence of preceding clips, capped by a maximum window length $L$. This design allows us to evaluate robustness under diverse context lengths.

\noindent\textbf{Hard Negative Construction.}
We formulate evaluation as a multiple-choice ranking task. For each query, we construct a candidate pool $\mathcal{C}$ containing the ground-truth clip and a set of mined negatives. To explicitly diagnose consistency failures, we partition negatives into three types:
\begin{itemize}
    \setlength{\itemsep}{0pt}
    \setlength{\parskip}{0pt}
    \item \textbf{State Negatives (Temporal Inconsistency):}
    Sampled from the \textit{same} video but from different non-target step segments. These clips preserve the environment and actor identity but correspond to an invalid procedural state, including both past and future steps.

    \item \textbf{Identity Negatives (Appearance Inconsistency):}
    Sampled from \textit{different} videos (excluding the source video) using dataset-specific semantic or structural matching rules. For YouCook2, we mine semantically similar captions using Sentence-BERT. For COIN and CrossTask, we use task and step annotations to construct semantically matched cross-video distractors. These negatives remain semantically relevant but violate identity consistency.

    \item \textbf{Easy Negatives:}
    Randomly sampled clips with low semantic similarity, used to maintain a fixed candidate-pool size.
\end{itemize}

\noindent\textbf{Formal Protocol.}
Formally, the candidate set is defined as $\mathcal{C} = \{v_{gt}\} \cup \mathcal{N}_{\text{state}} \cup \mathcal{N}_{\text{identity}} \cup \mathcal{N}_{\text{easy}}$, and performance is evaluated by ranking candidates within this pool. Unless otherwise specified, we adopt a fixed 1-vs-9 protocol, in which each query is evaluated against one ground-truth clip and nine negatives. We first sample up to 3 state negatives and up to 3 identity negatives. If one hard-negative pool is insufficient, the remaining slots are backfilled from the other hard-negative pool; any leftover slots are then filled with easy negatives, keeping the total candidate pool size fixed at 10. This composition is used only for evaluation; training-time candidate construction is described separately in Appendix~\ref{sec:appendix_cast_details}. By construction, semantic matching alone is insufficient: to identify the correct target, the model must satisfy both state and identity consistency. We report Recall@K within the candidate pool, and optionally provide results stratified by negative type. In this way, the protocol isolates consistency errors while controlling for semantic difficulty. Additional dataset-specific construction details are provided in \S\ref{sec:experiments} and the appendix.

\subsection{CAST: Context-Aware State Transition}
\label{subsec:cast_model}

Building on the formulation in \S\ref{subsec:task_definition} and the benchmark design in \S\ref{subsec:eval_protocol}, we propose \textbf{CAST}, a lightweight query-side adapter atop a frozen backbone that models procedural steps as state-conditioned transitions in the embedding space. In contrast to standard dual-encoders that statically match text to video, CAST explicitly predicts the next visual state through a residual update.

\noindent\textbf{Overview.}
CAST is implemented as a lightweight adapter on top of a frozen pre-trained video encoder (e.g., CLIP), while remaining plug-and-play at inference time. Given the anchor state $v_{t-1}$, the instruction $q_t$, and the visual history $\mathcal{H}_t$, the goal is to predict the target representation $\hat{v}_t$. We formulate this prediction as a residual transition:
\begin{equation}
    \hat{v}_t = \phi(v_{t-1} + \Delta(v_{t-1}, q_t, \mathcal{H}_t)),
    \label{eq:cast_objective}
\end{equation}
where $\Delta \in \mathbb{R}^d$ denotes the predicted transition vector and $\phi(\cdot)$ denotes L2 normalization.

\noindent\textbf{Why Simple Residual Modeling Works.} 
The central intuition behind CAST is that procedural actions (e.g., ``slice the tomato'') primarily alter state-related visual factors, while much of the scene, including identity and background, remains persistent. Modeling the transition $\Delta$ additively therefore introduces a useful inductive bias: the residual branch can focus on procedural change, while the anchor state $v_{t-1}$ carries persistent identity cues through the residual connection. The type-aware contrastive objective further encourages the predicted update to emphasize state-relevant changes rather than static appearance cues, echoing the intuition of content-motion decomposition~\cite{villegas2017decomposing} without additional complexity.

\noindent\textbf{Dual-Path Transition Prediction.}
To capture both local state change and broader narrative context, we decompose the transition vector into two complementary components, $\Delta = \Delta_{cond} + \Delta_{ctx}$.

\noindent\textit{1) Path 1: Instruction-Conditioned State Transition.}
In contrast to standard approaches that model actions purely from text, CAST grounds the instruction in the current visual state. Concretely, we concatenate the text embedding $f_t(q_t)$ with the anchor visual state $v_{t-1}$ and process the result through an MLP:
\begin{equation}
    \Delta_{cond} = \text{MLP}_{cond}([f_t(q_t); v_{t-1}]).
\end{equation}
This branch predicts how the specific scene encoded by $v_{t-1}$ should evolve under instruction $q_t$, thereby tightly coupling action semantics with the current observation.

\noindent\textit{2) Path 2: Temporal Context Attention.}
To resolve ambiguities that depend on longer-term history (e.g., distinguishing ``stir'' in early versus late cooking stages), we employ a multi-head cross-attention mechanism~\cite{vaswani2017attention}. The instruction embedding $f_t(q_t)$ serves as the \textit{query}, while the visual history sequence $\mathcal{H}_t$ provides the \textit{keys} and \textit{values}. The resulting attention output is projected by an MLP to obtain the context-dependent transition component:
\begin{equation}
\Delta_{ctx} = \text{MLP}_{ctx}\left( \text{MultiHead}(Q = f_t(q_t), K = \mathcal{H}_t, V = \mathcal{H}_t) \right)
\end{equation}
This branch aggregates relevant historical cues into a context-aware adjustment vector $\Delta_{ctx}$.

\noindent\textbf{State as an Emergent Property.}
Importantly, CAST does not assume a discrete or pre-defined state space. Instead, it treats visual state as an emergent property of the embedding geometry shaped by procedural supervision. By modeling $\Delta$ as an instruction-conditioned vector shift, rather than as a scalar reweighting of existing features, CAST predicts a target embedding that extrapolates the current state toward a plausible next step. This notion is operationalized through measurable gains in retrieval consistency under the controlled hard-negative settings of our CVR protocol, without requiring task-specific state annotations.

Detailed architectural specifications, including layer dimensions, attention configuration, normalization, and padding or masking strategy, are provided in Appendix~\ref{sec:appendix_cast_details}.

\subsection{Training and Inference}
\label{subsec:training_inference}

\paragraph{Type-Aware Contrastive Objective.}
We freeze the backbone and optimize only the CAST adapter. For each query step, CAST predicts the next-state embedding $\hat{v}_t$ from the instruction and visual history. Let $v_t^{+}$ denote the ground-truth continuation embedding, $\mathcal{N}_{s}(t)$ the mined state negatives, and $\mathcal{N}_{i}(t)$ the mined identity negatives. We use cosine similarity $\mathrm{sim}(\cdot,\cdot)$ with temperature $\tau$.

We begin with a standard in-batch InfoNCE objective for global discrimination:
\begin{equation}
\mathcal{L}_{\text{batch}}
=
-\frac{1}{B}\sum_{t=1}^{B}
\log
\frac{
\exp\!\left(\mathrm{sim}(\hat{v}_t, v_t^{+})/\tau\right)
}{
\sum_{j=1}^{B}
\exp\!\left(\mathrm{sim}(\hat{v}_t, v_j^{+})/\tau\right)
},
\end{equation}
where $B$ is the batch size.
To explicitly enforce fine-grained consistency beyond global batch discrimination, we further introduce two local contrastive objectives:
\begin{equation}
\mathcal{L}_{\text{state}}
=
-\frac{1}{B}\sum_{t=1}^{B}
\log
\frac{
\exp\!\left(\mathrm{sim}(\hat{v}_t, v_t^{+})/\tau\right)
}{
\exp\!\left(\mathrm{sim}(\hat{v}_t, v_t^{+})/\tau\right)
+
\sum_{v \in \mathcal{N}_{s}(t)}
\exp\!\left(\mathrm{sim}(\hat{v}_t, v)/\tau\right)
},
\end{equation}

\begin{equation}
\mathcal{L}_{\text{ident}}
=
-\frac{1}{B}\sum_{t=1}^{B}
\log
\frac{
\exp\!\left(\mathrm{sim}(\hat{v}_t, v_t^{+})/\tau\right)
}{
\exp\!\left(\mathrm{sim}(\hat{v}_t, v_t^{+})/\tau\right)
+
\sum_{v \in \mathcal{N}_{i}(t)}
\exp\!\left(\mathrm{sim}(\hat{v}_t, v)/\tau\right)
}.
\end{equation}

Rather than aggregating state and identity negatives into a single denominator, we optimize $\mathcal{L}_{\text{state}}$ and $\mathcal{L}_{\text{ident}}$ separately. This prevents highly overlapping identity negatives from dominating the gradients, thereby preserving a stronger signal for fine-grained state discrimination.

The final objective is:
\begin{equation}
\mathcal{L}
=
\mathcal{L}_{\text{batch}}
+
\lambda_{s}\mathcal{L}_{\text{state}}
+
\lambda_{i}\mathcal{L}_{\text{ident}}.
\end{equation}

Here, $\lambda_s > \lambda_i$ reflects our design prior that fine-grained state discrimination constitutes the primary challenge in CVR. At the same time, the residual prediction design anchors the transition around $v_{t-1}$, which already provides a useful bias toward retaining identity-consistent context.

\paragraph{Plug-and-Play Inference.}
At inference time, CAST operates as a scalable query-side plug-and-play module. The video gallery is pre-indexed once using the frozen backbone. Given a query step, CAST computes $\hat{v}_t$ on-the-fly from the visual history and instruction, and scores candidate continuations in the same frozen embedding space.

For a query instruction $q$ and a candidate clip $c$, we define three scores:
\begin{align}
A(q,c) &= \mathrm{sim}(t_q, v_c), \\
B(q,c) &= \mathrm{sim}(v_{t-1}, v_c), \\
C(q,c) &= \mathrm{sim}(\hat{v}_t, v_c),
\end{align}
where $t_q$ is the query text embedding, $v_{t-1}$ is the last context-clip embedding, $\hat{v}_t$ is the CAST-predicted next-state embedding, and $v_c$ is the candidate clip embedding.
Here, $A$ captures semantic alignment, $B$ visual continuity with the anchor state, and $C$ compatibility with the predicted future state.

We combine these signals using the \emph{Full Ensemble}:
\begin{equation}
S(q,c) = A(q,c) + w_v B(q,c) + w_p C(q,c).
\label{eq:full_ensemble}
\end{equation}
The ensemble coefficients are selected on a held-out validation split for each dataset and backbone setting, and then kept fixed during evaluation. The coefficient selection protocol and search ranges are provided in Appendix~\ref{sec:appendix_cast_details}.
Because CAST operates exclusively on the query side and does not require re-indexing the video gallery, it preserves the scalability of standard retrieval pipelines while enabling retrieval under temporal context.

%% file: sec/3_exp.tex
\section{Experiments}
\label{sec:experiments}

\subsection{Experimental Setup}
\label{subsec:setup}

\begin{table}[t]
    \centering
    \small
    \caption{\textbf{CVR benchmark statistics.}
We report the size of the final CVR evaluation set for each dataset.
All settings follow the fixed 1-vs-9 ranking protocol described in Sec.~\ref{subsec:eval_protocol}.
Dataset-specific negative mining rules are summarized in Appendix Table~\ref{tab:benchmark_mining}.}
    \label{tab:benchmark_stats}
    \vspace{-4pt}
    \setlength{\tabcolsep}{5pt}
    \begin{tabular}{l l c c c}
        \toprule
        \textbf{Dataset} & \textbf{Split Protocol} & \textbf{\#Videos} & \textbf{\#Step-Clips} & \textbf{\#Queries} \\
        \midrule
        YouCook2~\cite{zhou2018towards} & official train; CVR eval on val & 414 & 3,179 & 2,765 \\
        COIN~\cite{tang2019coin} & official train; CVR eval on test & 2,134 & 6,241 & 4,107 \\
        CrossTask~\cite{zhukov2019cross} & video-disjoint 80/20 split & 509 & 2,731 & 2,222 \\
        \bottomrule
    \end{tabular}
    \vspace{-6pt}
\end{table}
\paragraph{Datasets.}
We evaluate CAST on three procedural video datasets. \textbf{YouCook2}~\cite{zhou2018towards} contains $\sim$2k cooking videos across 89 recipes; we train on the official training split and construct the CVR evaluation benchmark exclusively from the official validation split. \textbf{COIN}~\cite{tang2019coin} comprises 11,827 videos spanning 180 tasks; we train on the official training split and construct the evaluation benchmark from the official test split. \textbf{CrossTask}~\cite{zhukov2019cross} (4.7k videos, 83 tasks) broadens the domain coverage by including both culinary and non-culinary activities, such as car maintenance. For CrossTask, we first construct the CVR benchmark from annotated videos and then split it at the video level into disjoint training and evaluation subsets (80\%/20\%, random seed 42), yielding \texttt{crosstask\_train.json} and \texttt{crosstask\_test.json}. In all cases, training and evaluation are video-disjoint, and benchmark queries and candidate pools are constructed exclusively from the held-out evaluation split. In the main CLIP-based experiments, each clip is represented by mean-pooled CLIP features extracted from 3 uniformly sampled frames. Backbone-specific feature extraction settings for the universality experiments are provided in Appendix~\ref{sec:appendix_backbone}.

\paragraph{Dataset Split Protocol.}
We explicitly separate weight training, ensemble-coefficient selection, and final benchmark evaluation. For YouCook2 and COIN, CAST is trained on the official training split, while final retrieval metrics are reported only on the held-out evaluation benchmark constructed from the official validation and test splits, respectively. For CrossTask, the benchmark is first constructed from annotated videos and then split by unique video ID into disjoint training and evaluation subsets, ensuring that no source video appears in both partitions. Any held-out validation subset used for selecting Full Ensemble coefficients is drawn exclusively from the training portion and is never used for final benchmark evaluation.

\paragraph{CVR Benchmark Protocol.}
To evaluate consistency, we adopt a fixed \textbf{1-vs-9 multiple-choice ranking} protocol. For each query step $q_t$, the candidate pool $\mathcal{C}$ has 10 items: (1) one ground-truth clip; (2) up to 3 \textbf{Hard State Negatives} sampled from the same video but at different time steps, thereby testing temporal causality while preserving scene and identity cues; (3) up to 3 \textbf{Hard Identity Negatives} sampled from different videos using dataset-specific semantic matching rules to test visual identity preservation under high semantic overlap; and (4) easy negatives used to fill the remaining slots and maintain a fixed 10-way ranking setting. If one hard-negative pool is insufficient, the remaining slots are first backfilled from the other available hard-negative pool, and any leftover slots are then filled with easy negatives. For YouCook2, identity negatives are mined via Sentence-BERT~\cite{reimers2019sentence} caption similarity, while for COIN and CrossTask we use task and step annotations to construct semantically matched cross-video distractors. The context history $\mathcal{H}_t$ contains up to the preceding $L=5$ clips. Detailed benchmark mining rules are provided in Appendix~\ref{sec:appendix_cast_details}.

\begin{table*}[t]
    \centering
    \small 
    \caption{\textbf{CVR Benchmark Results (CLIP-B/32).} CAST provides the most favorable overall trade-off across datasets and diagnostic metrics, with especially clear gains on state-sensitive retrieval. Diagnostic scores (State Acc. and Ident. Acc.) are averaged across the three benchmark datasets.}
    \label{tab:main_comparison}
    \setlength{\tabcolsep}{3.5pt}
    \begin{tabular}{l | c | cc | cc | cc | cc}
        \toprule
        \multirow{2}{*}{\textbf{Method}} & \textbf{Context} & \multicolumn{2}{c|}{\textbf{YouCook2}} & \multicolumn{2}{c|}{\textbf{COIN}} & \multicolumn{2}{c|}{\textbf{CrossTask}} & \multicolumn{2}{c}{\textbf{Diagnostic}} \\
        \cmidrule(lr){3-4} \cmidrule(lr){5-6} \cmidrule(lr){7-8} \cmidrule(lr){9-10}
         & \textbf{Modeling} & \textbf{Acc.}$\uparrow$ & \textbf{MnR}$\downarrow$ & \textbf{Acc.}$\uparrow$ & \textbf{MnR}$\downarrow$ & \textbf{Acc.}$\uparrow$ & \textbf{MnR}$\downarrow$ & \textbf{State}$\uparrow$ & \textbf{Ident.}$\uparrow$ \\
        \midrule
        CLIP Baseline & Context-Free & 25.03 & 3.60 & 14.10 & 3.91 & 16.83 & 4.15 & 45.52 & 28.90 \\
        Late Fusion (Heuristic) & Fixed Weighting & 31.10 & 2.56 & 17.85 & 3.28 & 22.05 & 2.86 & 28.69 & 68.29 \\
        Late Fusion (Learned)   & Learned Weighting & 36.60    & 2.53    & \textbf{44.66}    & \textbf{2.11}    & 25.52    & 2.86    & 40.06    & 76.06    \\
        Early Fusion  & Feature Concat. & 35.99 & 2.28 & 15.12 & 2.60 & 35.29 & 2.36 & 31.14 & \textbf{83.59} \\
        \midrule
        \rowcolor{gray!10} \textbf{CAST (Ours)} & \textbf{State Transition} & \textbf{44.77} & \textbf{2.15} & 40.47 & 2.16 & \textbf{47.39} & \textbf{2.14} & \textbf{53.81} & 74.67 \\
        \bottomrule
    \end{tabular}
\end{table*}
\paragraph{Baselines.}
We compare CAST against four baselines: (i) \textbf{Zero-Shot Baseline}, a context-free matching formulation using the frozen backbone; (ii) \textbf{Late Fusion (Heuristic)}, a fixed weighted sum of semantic and visual scores ($S = sim(q, v) + \alpha \cdot sim(v_{t-1}, v)$); (iii) \textbf{Late Fusion (Learned)}, a lightweight MLP trained to learn dataset-specific weighting over semantic and visual scores, serving as a strong aggregation baseline; and (iv) \textbf{Early Fusion}, a trainable MLP that concatenates text and context features before prediction.

\paragraph{Implementation Details.}
We use a frozen CLIP (ViT-B/32)~\cite{radford2021learning} backbone. CAST is trained with AdamW~\cite{loshchilov2017decoupled} ($lr=1e^{-4}$, weight decay=$1e^{-3}$) for 30, 20, and 50 epochs on YouCook2, COIN, and CrossTask, respectively, with batch size 512. To emphasize temporal state discrimination, we set the loss weights to $\lambda_s=5.0$ and $\lambda_i=1.0$. The context window is $L=5$. Unless otherwise specified, the reported CAST performance uses the Full Ensemble score defined in Eq.~\eqref{eq:full_ensemble}, with ensemble coefficients selected on a held-out validation subset for each setting and then kept fixed at evaluation time. Details on training splits, hard-negative sampling, backbone-specific feature extraction, and inference settings are provided in Appendix~\ref{sec:appendix_cast_details} and Appendix~\ref{sec:appendix_backbone}.

\paragraph{Metrics.}
We report \textbf{Accuracy (Acc.)} (Recall@1) and \textbf{Mean Rank (MnR)}. To diagnose failure modes more precisely, we additionally define \textbf{State Acc.} (the ground-truth clip ranked above all State Negatives) and \textbf{Ident. Acc.} (the ground-truth clip ranked above all Identity Negatives). For a global assessment of state discrimination and identity preservation, the diagnostic scores reported in our main results (Table~\ref{tab:main_comparison} and Table~\ref{tab:universality}) are averaged across the three benchmark datasets.

\noindent \textbf{Control for Semantic Difficulty.} Importantly, our 1-vs-9 ranking protocol is designed so that performance gains cannot be attributed solely to vision-language alignment. By constructing each candidate pool from a mixture of state, identity, and easy negatives, we require the model to move beyond keyword matching and leverage the visual context $\mathcal{H}_t$ to resolve temporal ambiguities. This controlled setup makes the improvements reported in Table~\ref{tab:main_comparison} more directly interpretable as evidence of stronger state discrimination and more reliable identity preservation under contextual constraints.

\subsection{Quantitative Results on CVR Benchmark}
\label{subsec:main_results}

We organize our quantitative analysis around two central questions: (1) Does explicit state-transition modeling outperform standard context aggregation? (2) Does CAST transfer effectively across diverse frozen foundation-model backbones?

\noindent\textbf{1. Effectiveness of the CAST Mechanism.}
In Table~\ref{tab:main_comparison}, we fix the backbone to CLIP (ViT-B/32) in order to isolate the contribution of the adapter architecture itself.

\textbf{Limitations of Scalar Aggregation.}
To test whether simple score re-weighting is sufficient, we include a strong Late Fusion (Learned) baseline. On COIN, where consecutive procedural steps often exhibit strong visual similarity, the learned fusion can exploit visual inertia~\cite{geirhos2020shortcut}, achieving its best performance (44.66\%) and slightly surpassing CAST (40.47\%). However, this shortcut does not generalize to datasets with more substantial state transitions, such as CrossTask and YouCook2, where procedural steps involve more pronounced visual changes (e.g., ``add ginger paste'' $\rightarrow$ ``fry''). In these settings, the learned baseline degrades sharply, dropping to 25.52\% on CrossTask, which is 21.9 points below CAST in accuracy.

\textbf{Vector Transition vs. Scalar Weighting.} 
The cross-dataset performance gap highlights a fundamental difference between CAST and aggregation-based baselines. Whereas scalar late-fusion models can only reweight existing similarity signals, CAST predicts a structured, instance-conditioned residual $\Delta$ in the latent space. This formulation is more effective at resolving causal ambiguities, as evidenced by CAST's consistently stronger State Accuracy under the diagnostic protocol.

\begin{table*}[t]
    \centering
    \small 
    \caption{\textbf{Universality Across Backbones.} CAST consistently improves over the corresponding zero-shot baseline across diverse frozen video and multimodal embedding models, with both methods operating in the same frozen native vision-language embedding space for each backbone. Diagnostic scores are averaged across the three benchmark datasets.}
    \label{tab:universality}
    \setlength{\tabcolsep}{3pt} %
    \begin{tabular}{l | c | cc | cc | cc | cc}
        \toprule
        \multirow{2}{*}{\textbf{Backbone}} & \multirow{2}{*}{\textbf{Setting}} & \multicolumn{2}{c|}{\textbf{YouCook2}} & \multicolumn{2}{c|}{\textbf{COIN}} & \multicolumn{2}{c|}{\textbf{CrossTask}} & \multicolumn{2}{c}{\textbf{Diagnostic}} \\
        \cmidrule(lr){3-4} \cmidrule(lr){5-6} \cmidrule(lr){7-8} \cmidrule(lr){9-10}
         & & \textbf{Acc.} & \textbf{MnR} & \textbf{Acc.} & \textbf{MnR} & \textbf{Acc.} & \textbf{MnR} & \textbf{State} & \textbf{Ident.} \\
        \midrule
        \rowcolor{gray!10} \multicolumn{10}{l}{\textit{Category I: Video Foundation Models}} \\
        \multirow{2}{*}{InternVideo2-1B~\cite{wang2024internvideo2}} 
          & Zero-Shot & 36.75 & 2.59 & 17.99 & 3.36 & 20.61 & 3.31 & 65.70 & 30.85 \\ 
          & \textbf{+ CAST} & \textbf{71.68} & \textbf{1.48} & \textbf{51.03} & \textbf{1.90} & \textbf{64.36} & \textbf{1.71} & \textbf{75.43} & \textbf{77.77} \\
        \midrule
        \multirow{2}{*}{VideoPrism-B~\cite{zhao2024videoprism}} 
          & Zero-Shot & 47.45 & 2.13 & 17.60 & 3.32 & 20.25 & 3.24 & 68.38 & 33.68 \\
          & \textbf{+ CAST} & \textbf{75.59} & \textbf{1.38} & \textbf{51.64} & \textbf{1.90} & \textbf{62.11} & \textbf{1.74} & \textbf{76.92} & \textbf{77.66} \\
        \midrule
        \rowcolor{gray!10} \multicolumn{10}{l}{\textit{Category II: Multimodal Embedding Models}} \\
        \multirow{2}{*}{GME-Qwen2-VL-2B~\cite{zhang2024gme}} 
          & Zero-Shot & 29.62 & 3.10 & 17.17 & 3.44 & 19.40 & 3.61 & 56.31 & 29.73 \\
          & \textbf{+ CAST} & \textbf{54.39} & \textbf{1.95} & \textbf{45.68} & \textbf{2.05} & \textbf{52.43} & \textbf{2.04} & \textbf{67.20} & \textbf{72.28} \\
        \midrule
        \multirow{2}{*}{Qwen3-VL-Embedding-2B~\cite{qwen3vlembedding}} 
          & Zero-Shot & 33.45 & 2.89 & 17.73 & 3.50 & 19.44 & 3.56 & 58.44 & 29.79 \\
          & \textbf{+ CAST} & \textbf{56.64} & \textbf{1.85} & \textbf{44.87} & \textbf{2.09} & \textbf{48.96} & \textbf{2.09} & \textbf{66.18} & \textbf{69.11} \\
        \bottomrule
    \end{tabular}
\end{table*}
\noindent\textbf{2. Universality Across Backbones.}
While Table~\ref{tab:main_comparison} isolates the CAST mechanism under a fixed CLIP (ViT-B/32) backbone, Table~\ref{tab:universality} evaluates whether the same design transfers across diverse \emph{frozen} video and multimodal embedding models. To assess robustness across architectures, we apply CAST atop two categories of SOTA foundation models:
\begin{itemize}
        \item \textbf{Video Foundation Models}: models specialized for temporal dynamics, including \textbf{InternVideo2}~\cite{wang2024internvideo2} and \textbf{VideoPrism}~\cite{zhao2024videoprism}.
    \item \textbf{Multimodal Embedding Models}: unified retrieval-oriented models, including the General Multimodal Embedding framework \textbf{GME-Qwen2-VL-2B}~\cite{zhang2024gme} and the state-of-the-art \textbf{Qwen3-VL-Embedding}~\cite{qwen3vlembedding}.
\end{itemize}
For each backbone, we keep the native text and video encoders frozen and operate within its original vision-language embedding space. Concretely, the zero-shot baseline in Table~\ref{tab:universality} uses the backbone's native vision-language similarity, while CAST is trained in the same frozen space using backbone-specific text, context-clip, and candidate-video embeddings. The CAST inference scores are defined within each backbone space: the semantic score $A$ is the native vision-language similarity, the visual continuity score $B$ is the similarity between the last context clip and candidate clip, and the prediction score $C$ is the similarity between the CAST-predicted next-state embedding and the candidate clip. Universality is therefore evaluated by changing only the frozen backbone-specific embedding space, while keeping the CAST architecture, training objective, and evaluation protocol unchanged.

\noindent The gains are consistent across all backbone families. On InternVideo2, CAST nearly doubles the zero-shot accuracy on YouCook2 ($36.75\% \rightarrow 71.68\%$) and more than triples it on CrossTask ($20.61\% \rightarrow 64.36\%$). Similarly, under VideoPrism, CAST yields an absolute gain of $+28.1\%$ on YouCook2. The same pattern holds for multimodal embedding models: GME-Qwen2-VL-2B improves by $+24.6\%$ on YouCook2, while Qwen3-VL-Embedding improves from $33.45\%$ to $56.64\%$. Backbone checkpoints, text and video preprocessing, and feature extraction protocols are provided in Appendix~\ref{sec:appendix_backbone}.

\noindent\textbf{Identity Consistency and Embedding Quality.}
Across all backbones, CAST consistently improves identity preservation under hard identity negatives, raising Ident. accuracy from roughly 30\% to 69--78\% (Table~\ref{tab:universality}). Notably, the gains scale with the quality of the underlying embedding space, suggesting that the same CAST design transfers effectively across different frozen spaces while more fully exploiting stronger visual representations.

\begin{figure*}[t]
    \centering
    \includegraphics[width=\linewidth, trim={0 0.6cm 0 0.6cm}, clip]{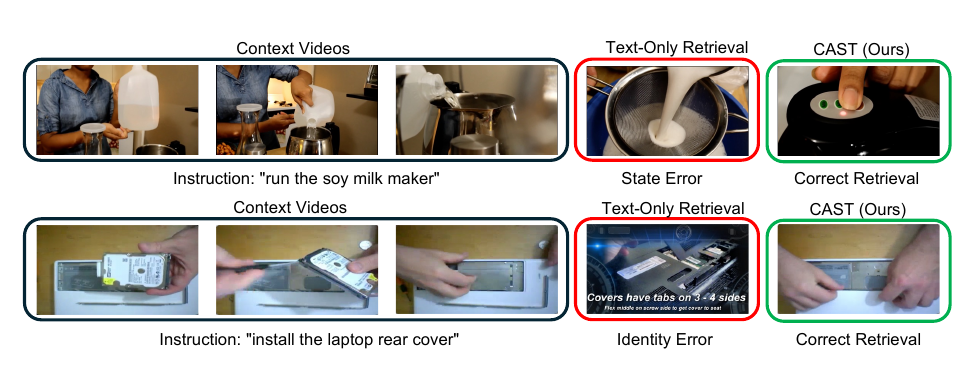}
    \caption{\textbf{Qualitative retrieval examples on the CVR benchmark.}
Given the same procedural context and instruction, context-agnostic retrieval often returns semantically relevant but temporally inconsistent clips, producing either a \textit{State Error} or an \textit{Identity Error}. By contrast, CAST retrieves the correct continuation in both cases by modeling state transitions conditioned on visual history.}
    \label{fig:retrieval_qual_examples}
    \vspace{-4pt}
\end{figure*}

\noindent\textbf{Qualitative Retrieval Examples.}
To complement the metrics, Figure~\ref{fig:retrieval_qual_examples} presents two representative CVR retrieval cases. In both examples, the context-agnostic baseline retrieves a semantically related clip but fails to preserve either the current procedural state or the visual identity. By contrast, CAST retrieves the correct continuation by explicitly modeling the underlying state transition.
\subsection{Ablation Studies}
\label{subsec:ablation}

To isolate the sources of CAST's gains, we conduct ablation studies over both the architecture design (Table~\ref{tab:ablation_arch}) and the inference formulation (Table~\ref{tab:ablation_inf}).

\noindent\textbf{1. Architecture and Objective Analysis.}
Table~\ref{tab:ablation_arch} presents a factorial analysis of our design choices on the YouCook2 validation benchmark. First, at the objective level, comparing the first two rows shows that \textit{Residual Modeling} ($\Delta$) is critical. Replacing direct target prediction with an additive residual update improves the Early Fusion model from 35.99\% to 38.95\%, while raising State Accuracy from 38.92\% to 43.51\%. This result supports the central inductive bias of CAST: anchoring prediction around $v_{t-1}$ makes procedural continuation easier to model, because the adapter only needs to predict a targeted state update rather than reconstruct the full next-step embedding from scratch.

Second, at the architectural level, comparing \textit{Early Fusion + Residual} with CAST (Dual-Path + Residual) reveals a gain of +4.05\% Acc. This indicates that simple feature concatenation is insufficient for modeling noisy procedural context. By contrast, CAST's dual-path design and cross-attention mechanism provide a more effective way to isolate temporally relevant cues from visual history.

\begin{table}[t]
    \centering
    \small 
    \caption{\textbf{Ablation of target formulation and architecture on the YouCook2 validation benchmark.}
Residual modeling consistently outperforms direct target prediction, and the dual-path CAST design further improves over simple early fusion.}
    \label{tab:ablation_arch}
    \setlength{\tabcolsep}{3pt} 
    \begin{tabular}{lc|c|cc}
        \toprule
        \textbf{Fusion} & \textbf{Target} & \textbf{Acc.} & \textbf{State} & \textbf{Ident.} \\
        \midrule
        Early (Concat) & Direct $(\hat{v}_t)$ & 35.99 & 38.92 & \textbf{83.58} \\
        Early (Concat) & Residual $(\Delta)$ & 38.95 & 43.51 & 81.99 \\
        \midrule
        \rowcolor{gray!10} \textbf{CAST (Ours)} & \textbf{Residual $(\Delta)$} & \textbf{44.77} & \textbf{51.03} & 78.48 \\
        \bottomrule
    \end{tabular}
\end{table}

\begin{table}[t]
    \centering
    \small 
    \caption{\textbf{Inference signal decomposition on the YouCook2 validation benchmark.}
We analyze the three inference components defined in Sec.~\ref{subsec:training_inference} and their combinations. While the semantic ensemble achieves the highest exact-match accuracy, the full ensemble provides a more balanced trade-off between state discrimination and identity preservation.}
    \label{tab:ablation_inf}
    \setlength{\tabcolsep}{8pt} 
    \begin{tabular}{l | c | cc}
        \toprule
        \textbf{Inference Strategy} & \textbf{Acc.} & \textbf{State} & \textbf{Ident.} \\
        \midrule
        A. Text Matching ($q$) & 25.03 & 50.45 & 25.32 \\ 
        B. Vis. Continuity ($v_{t-1}$) & 25.90 & 27.70 & \textbf{81.95} \\ 
        C. CAST Prediction ($\hat{v}_t$) & 42.60 & 50.81 & 75.99 \\
        \midrule
        \textit{Semantic Ens. (A+C)} & \textbf{45.46} & \textbf{56.71} & 70.38 \\
        \rowcolor{gray!10} \textbf{Full Ens. (A+B+C)} & 44.77 & 51.03 & 78.48 \\
        \bottomrule
    \end{tabular}
\end{table}
\noindent\textbf{2. Inference Components.}
Table~\ref{tab:ablation_inf} decomposes the three inference components defined in Sec.~\ref{subsec:training_inference}: the semantic matching score $A$, the visual continuity score $B$, and the CAST prediction score $C$, together with their \emph{Full Ensemble} combination in Eq.~\eqref{eq:full_ensemble}.

\begin{figure}[t]
    \centering
    \includegraphics[width=0.4\linewidth]{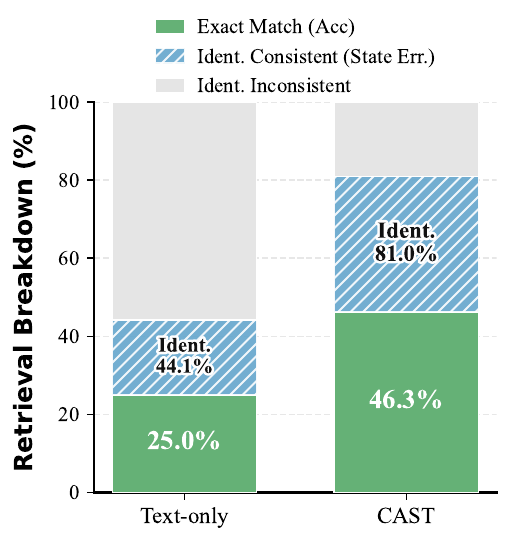}
    \caption{\textbf{Retrieval quality breakdown.}
We categorize top-1 retrieval outcomes as \textit{Exact Match} (green), \textit{Identity Consistent but State-misaligned} (blue), or \textit{Identity Inconsistent} (gray). CAST yields identity-consistent outcomes in 81.0\% of queries (green+blue), substantially improving over the text-only baseline. This differs from the Ident.\ Acc.\ metric in Table~\ref{tab:ablation_inf}, which requires the ground-truth clip to rank above all identity negatives.}
    \label{fig:retrieval_breakdown}
    \vspace{-6pt}
\end{figure}

Several trends are apparent. Semantic matching alone improves recall but severely degrades identity preservation. Among the individual components, CAST prediction ($C$) already provides the most balanced performance, indicating that the predicted future-state embedding captures a meaningful procedural continuation signal. While the Semantic Ensemble ($A{+}C$) attains the highest overall accuracy, adding visual continuity ($B$) improves identity preservation from 70.38 to 78.48, with only a modest drop in exact-match accuracy. We therefore adopt the Full Ensemble as the default inference strategy, since it provides the most balanced overall trade-off between state discrimination and identity preservation. Figure~\ref{fig:retrieval_breakdown} provides a qualitative breakdown of this trade-off.

\begin{figure}[t]
    \centering
    \includegraphics[width=0.95\linewidth]{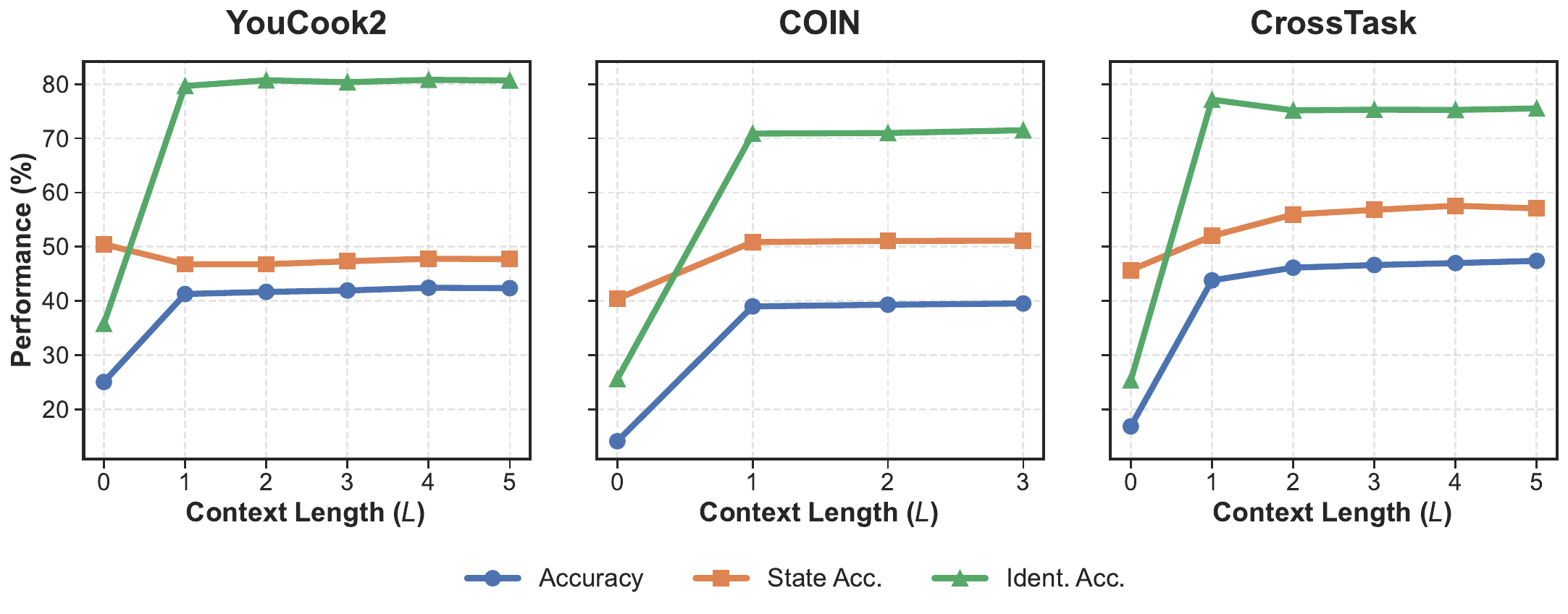}
    \caption{\textbf{Effect of context history length ($L$).}
Results are shown on YouCook2, COIN, and CrossTask. Performance improves sharply from $L=0$ to $L=1$, and then largely saturates as $L$ increases further.}
    \label{fig:ablation_context}
\end{figure}

\noindent\textbf{3. Context History Length.} 
Figure~\ref{fig:ablation_context} illustrates the impact of the context window length $L$ on model performance across YouCook2, COIN, and CrossTask. Across all three datasets, the largest gain occurs when moving from the context-agnostic setting ($L=0$) to using only the immediate predecessor ($L=1$). In particular, both Accuracy and Ident.\ Acc.\ improve sharply at $L=1$, highlighting that even minimal visual history provides a strong cue for resolving procedural continuity.

As $L$ increases beyond 1, the gains largely saturate, with only modest additional improvements or small fluctuations across datasets and metrics. This suggests that the immediate predecessor $v_{t-1}$ serves as the primary causal anchor, while longer history provides only diminishing returns.

\subsection{Application: Guiding Video Generation}
\label{subsec:generation_app}

Beyond retrieval, CAST can also rerank black-box video generation candidates using its predicted transition compatibility. Given a single visual context image, extracted as the last frame of the most recent context clip, together with a next-step instruction, we use Veo~\cite{veo2024} to generate $K=4$ candidate continuation videos for each prompt. We then compare two ranking strategies over the same candidate pool: \textbf{Standard Text Match}, which uses a text-only retrieval score, and \textbf{CAST Reranking}, which applies the same final reranking strategy used in retrieval by combining semantic matching, visual continuity, and CAST-predicted next-state compatibility.

To evaluate whether CAST improves candidate selection for black-box video generation, we conduct a blind human study on 300 prompts randomly sampled from the YouCook2 validation benchmark using a fixed random seed. No training-set queries are used in either the generation or human-evaluation protocol. For each prompt, standard text matching and CAST reranking each select one top-ranked video from the same set of generated candidates. Among the 300 prompts, 142 are overlap cases in which both methods select the same candidate (47.3\%), leaving 158 non-overlap prompts (52.7\%) for pairwise human evaluation. Equivalently, CAST changes the final selected candidate on 158/300 prompts, which defines the subset on which pairwise comparison is informative. We compare the two selected outputs in randomized order using three human annotators, and aggregate the final label for each prompt by majority vote. As shown in Table~\ref{tab:generation}, CAST-selected outputs are consistently preferred over standard text-only selection across all three dimensions: overall preference, physical plausibility, and temporal logic. These results provide preliminary evidence that the learned state transition serves as a useful reranking signal for black-box procedural video generation. Additional generation setup, protocol details, and qualitative examples are provided in Appendix~\ref{sec:qualitative_appendix}.

\begin{table}[t]
    \centering
    \small
    \caption{\textbf{Guiding Video Generation.} Blind human evaluation on 300 prompts from the YouCook2 validation benchmark. For each prompt, Veo generates $K=4$ candidate videos, and we compare the top-ranked outputs selected by standard text matching and CAST reranking. Of the 300 prompts, 158 are non-overlap cases where the two methods select different candidates; we report majority-vote preference rates from annotators on this subset.}
    \label{tab:generation}
    \setlength{\tabcolsep}{3pt}
    \begin{tabular}{l | >{\centering\arraybackslash}p{1.9cm} >{\centering\arraybackslash}p{2.2cm} >{\centering\arraybackslash}p{1.8cm}}
        \toprule
        \textbf{Selection Method} & \textbf{Overall Preference} & \textbf{Physical Plausibility} & \textbf{Temporal Logic} \\
        \midrule
        Standard Text Match & 38.6\% & 39.9\% & 38.6\% \\
        \rowcolor{gray!10} \textbf{CAST Reranking} & \textbf{55.1\%} & \textbf{50.6\%} & \textbf{52.5\%} \\
        Human Tie & 6.3\% & 9.5\% & 8.9\% \\
        \bottomrule
    \end{tabular}
\end{table}

%% file: sec/4_related_work.tex
\section{Related Work}
\label{sec:related_work}

\paragraph{Video-Text Retrieval.}
The dominant paradigm in video-text retrieval relies on dual-encoder architectures~\cite{radford2021learning} or generative-contrastive learners such as VideoCoCa~\cite{yan2022videococa}, which map clips and queries into a shared embedding space~\cite{bain2021frozen, luo2021clip4clip, wang2022internvideo, zhao2024videoprism}. Early efforts toward temporal modeling, such as VideoBERT~\cite{sun2019videobert}, relied on discretized visual tokens and were limited by vocabulary granularity. Subsequent temporal transformers, including HERO~\cite{li2020hero}, UniVL~\cite{luo2020univl}, and VIOLET~\cite{fu2021violet}, incorporate richer cross-modal attention mechanisms; however, they still predominantly optimize \textit{global} video-text alignment. By contrast, our work focuses on retrieval under procedural context, using an extrapolated visual state as the retrieval anchor to enforce causal consistency beyond standard feature matching.

\paragraph{Procedural Video Understanding.}
Goal-oriented procedures require modeling causal dependencies across steps~\cite{zhou2018towards, tang2019coin, zhukov2019cross}. Prior work has explored dense captioning~\cite{wang2018bidirectional}, procedural planning~\cite{chang2020procedure}, and visual state modeling, including state discovery in image collections and state-change detection in videos~\cite{isola2015discovering, souvcek2022look}. However, these approaches typically represent state change using discrete categories or localized cues, limiting their ability to distinguish visually similar yet temporally inconsistent states. In contrast, CAST targets \textit{Consistent Video Retrieval} (CVR), which requires a continuous transition model capable of discriminating between fine-grained but temporally incoherent states (e.g., ``pre-slicing'' vs.\ ``post-slicing'').

\paragraph{Connection to World Models.} 
Recent predictive world models~\cite{bardes2024revisiting, assran2025v} learn representations by forecasting future states in latent space. Unlike these largely unconditional predictive formulations, CAST addresses \textit{instruction-conditioned retrieval} by explicitly mapping human intent, encoded as query text, to visual state transitions. Following principles of representation learning~\cite{bengio2013representation}, we treat procedural state as an \textit{emergent geometric property} of the embedding space rather than a pre-defined semantic label. This formulation biases CAST toward modeling task-relevant progression while reducing reliance on static visual content and dataset-specific shortcuts~\cite{geirhos2020shortcut}, such as visual inertia.

%% file: sec/5_limit.tex
\section{Limitations}
\label{sec:limitations}

CAST remains subject to several limitations. First, the current context window is fixed ($L=5$). While sufficient for short procedural sequences, it remains limited in capturing longer-range dependencies, such as objects or state changes introduced several steps earlier. Second, as a lightweight adapter, CAST is inherently constrained by the representational quality of the frozen backbone. When the base encoder fails to resolve subtle state differences, such as fine-grained texture variations or object-configuration changes, CAST is correspondingly limited. Third, the residual transition is learned without explicit geometric constraints on $\Delta$. Although this formulation is effective empirically, it does not explicitly constrain the latent space to separate temporal progression from persistent identity cues. Promising directions for future work include hierarchical memory for longer-horizon reasoning and more structured regularization for state-transition modeling.

%% file: sec/6_conclusion.tex
\section{Conclusion}
We introduced \textbf{Consistent Video Retrieval (CVR)} to address a structural limitation of standard retrieval systems in maintaining temporal and identity coherence. We proposed CAST, a lightweight adapter that models procedural steps as state-conditioned residual transitions. Extensive experiments on YouCook2, COIN, and CrossTask show that CAST improves retrieval under procedural context, yielding clear gains on YouCook2 and CrossTask, remaining competitive on COIN, and consistently improving over the corresponding zero-shot baselines across diverse frozen embedding backbones. Beyond retrieval, we show that CAST can rerank candidate videos generated by black-box video generation models toward more coherent continuations. We hope this work encourages further research on context-aware inference and causally consistent video understanding.

%% file: sec/appendix.tex
\onecolumn
\section{Appendix: Backbone Details for Universality Experiments}
\label{sec:appendix_backbone}

This appendix documents the frozen backbones used in the
\emph{Universality Across Backbones} experiments, together with the exact
feature extraction and preprocessing protocols, to facilitate reproducibility.

\paragraph{General Protocol.}
For all backbone variants, we use publicly available pretrained checkpoints and
follow the authors' official inference or feature extraction pipelines.
All backbones remain frozen throughout, and no task-specific fine-tuning is applied.
For the universality results in Table~\ref{tab:universality}, each backbone operates in its own native frozen text-video embedding space.
Accordingly, both the zero-shot baseline and CAST use the backbone's native text encoder for queries and native video encoder for clips.
CAST retains the same architecture, training objective, and evaluation protocol as in the CLIP-based experiments, while operating in the corresponding backbone-specific frozen text-video embedding space.

Training schedules depend on both dataset and backbone.
For the main CLIP-based experiments, we train for 30 / 20 / 50 epochs on YouCook2 / COIN / CrossTask, respectively.
For the alternative frozen backbones (InternVideo2, VideoPrism, GME-Qwen2-VL-2B, and Qwen3-VL-Embedding-2B),
we train for 30 / 30 / 50 epochs on YouCook2 / COIN / CrossTask, respectively.

Frame sampling is likewise backbone-dependent.
CLIP, GME-Qwen2-VL-2B, and Qwen3-VL-Embedding-2B use 3 frames per clip on all datasets.
InternVideo2 uses 8 frames per clip on all datasets.
VideoPrism uses 8 frames per clip on YouCook2 and 4 frames per clip on COIN and CrossTask.
Frame-level features are aggregated into clip-level embeddings via mean pooling.
All extracted clip embeddings are L2-normalized prior to retrieval and CAST training.

\paragraph{Text Side and Score Definitions in Universality Experiments.}
For the universality results in Table~\ref{tab:universality}, both the zero-shot baseline and CAST are evaluated within each backbone's native frozen text-video embedding space.
For video foundation models such as InternVideo2 and VideoPrism, query text embeddings are obtained from the backbone's own frozen text encoder, and clip embeddings are obtained from the corresponding frozen video encoder.
For multimodal embedding backbones (e.g., GME-Qwen2-VL-2B and Qwen3-VL-Embedding-2B), both text and video embeddings are produced directly by the same pretrained multimodal embedding model.

Accordingly, the three inference scores are all defined within the corresponding backbone space.
The semantic score $A$ denotes the native text-video similarity between the query text embedding and a candidate clip embedding.
The visual continuity score $B$ denotes the similarity between the last context clip embedding and the candidate clip embedding.
The prediction score $C$ denotes the similarity between the CAST-predicted next-state embedding and the candidate clip embedding.
Thus, Table~\ref{tab:universality} compares zero-shot retrieval and CAST reranking within the same frozen backbone-specific embedding space, rather than under a shared text encoder.

\paragraph{InternVideo2 (PyTorch).}
We use the official InternVideo2 Stage-2 inference code and pretrained checkpoint.
Each clip is represented by 8 uniformly sampled frames.
Frames are resized to $224 \times 224$, normalized using ImageNet mean/std,
and packed into a tensor of shape $[B,T,C,H,W]$.
We extract clip embeddings using the model's video feature head (e.g., \texttt{get\_vid\_feat}),
keep the backbone frozen, and store one embedding per clip.
For efficiency, we perform extraction with multi-GPU data parallelism, shard the video list by rank,
save per-rank feature dictionaries, and merge them offline.

\paragraph{VideoPrism (JAX/Flax).}
We use the public VideoPrism LVT checkpoint together with the official Flax model and weight loader.
Each clip is represented by 8 uniformly sampled frames on YouCook2 and 4 uniformly sampled frames on COIN and CrossTask, with all frames resized to $288 \times 288$.
Pixels are scaled to $[0,1]$ and provided in NHWC format, as required by the JAX implementation.
Since the public VideoPrism forward interface expects text inputs, we pass dummy text IDs and paddings during video feature extraction and retain only the video embeddings.
For Table~\ref{tab:universality}, query text embeddings are still obtained from the backbone's native frozen text encoder; the dummy text inputs are used solely to satisfy the public video inference interface.
We extract one clip embedding per clip and L2-normalize the resulting vectors for retrieval.

\paragraph{Multimodal Embedding Models (GME-Qwen2-VL-2B, Qwen3-VL-Embedding).}
For multimodal embedding backbones, we use publicly available pretrained checkpoints
and operate directly on their embedding outputs without fine-tuning.
Each clip is represented by 3 uniformly sampled frames decoded with OpenCV and converted to PIL images.
We encode each frame with the embedding model to obtain frame-level embeddings, then mean-pool across frames
to form a clip embedding, followed by L2 normalization.

\paragraph{Qwen3-VL-Embedding.}
We use \texttt{Qwen3-VL-Embedding-2B} and extract embeddings via a lightweight wrapper
(\texttt{Qwen3VLEmbedder}). For efficiency, we run inference in FP16
(\texttt{torch\_dtype=float16}) with FlashAttention-2 (\texttt{attn\_implementation=flash\_attention\_2}).
Given a batch of clips, we flatten the 3 frames per clip, compute embeddings for all frames,
reshape them back to $[B,3,D]$, mean-pool over frames to obtain $[B,D]$ clip embeddings,
and apply L2 normalization before saving.
If decoding fails or a video is corrupted, we pad missing frames with black images to keep the input shape fixed.

\begin{table}[t]
\centering
\small
\caption{Backbone feature extraction settings for the universality experiments.
All backbones are frozen. For each backbone, zero-shot retrieval and CAST are both evaluated in the backbone's native text-video embedding space, while the table below summarizes the corresponding video-side feature extraction settings.}
\label{tab:appendix_backbone_settings}
\vspace{-4pt}
\setlength{\tabcolsep}{5pt}
\begin{tabular}{l|c|c|c|c}
\toprule
\textbf{Backbone} & \textbf{Impl.} & \textbf{\#Frames/clip (YC2 / COIN / CT)} & \textbf{Res.} & \textbf{Post-proc.} \\
\midrule
InternVideo2-1B & PyTorch & 8 / 8 / 8 & 224 & mean pool + L2 norm \\
VideoPrism-B & JAX/Flax & 8 / 4 / 4 & 288 & mean pool + L2 norm \\
GME-Qwen2-VL-2B & PyTorch & 3 / 3 / 3 & native & mean pool + L2 norm \\
Qwen3-VL-Embedding-2B & PyTorch & 3 / 3 / 3 & native & mean pool + L2 norm \\
\bottomrule
\end{tabular}
\vspace{-6pt}
\end{table}

\section{Appendix: CAST Training and Inference Details}
\label{sec:appendix_cast_details}

\paragraph{Dataset Split and Evaluation Protocol.}
We explicitly distinguish three roles for the data in all experiments: (1) the training split used to optimize CAST parameters; (2) a held-out validation subset drawn exclusively from the training portion, used to select the Full Ensemble coefficients; and (3) the final benchmark split used for all reported retrieval metrics.

For YouCook2, training instances are constructed from the official training split, whereas the fixed 1-vs-9 evaluation benchmark is constructed exclusively from the official validation split.
For COIN, training instances are constructed from the official training split, whereas the fixed evaluation benchmark is constructed from the official test split.
For CrossTask, we first construct a benchmark from the annotated videos and then partition it by unique video ID into disjoint training and evaluation subsets using an 80\%/20\% split with random seed 42, yielding \texttt{crosstask\_train.json} and \texttt{crosstask\_test.json}.
All reported CrossTask retrieval metrics are computed only on \texttt{crosstask\_test.json}, while training uses only \texttt{crosstask\_train.json}.
No source video appears in both training and evaluation.

\begin{table}[h]
\centering
\small
\caption{\textbf{Dataset split protocol used in CAST.}}
\label{tab:split_protocol}
\begin{tabular}{lccc}
\toprule
Dataset & Weight training & Coefficient tuning & Final benchmark eval \\
\midrule
YouCook2 & official train & held-out subset of train & official validation \\
COIN & official train & held-out subset of train & official test \\
CrossTask & \texttt{crosstask\_train.json} & held-out subset of train & \texttt{crosstask\_test.json} \\
\bottomrule
\end{tabular}
\end{table}

\paragraph{Benchmark Mining Details.}
The evaluation benchmark follows a fixed 1-vs-9 protocol across all datasets, consisting of 1 ground-truth clip and 9 negatives in total. For each query, we first sample up to 3 state negatives and up to 3 identity negatives. If one hard-negative pool is insufficient, we backfill the remaining slots from the other available hard-negative pool. Any remaining slots are then filled with easy negatives to preserve a fixed 9-negative candidate set. We always exclude the ground-truth clip itself from the negative pool, enforce unique candidate IDs within each query, and randomly shuffle the final candidate order to eliminate position bias.

\textbf{YouCook2.}
We construct the benchmark exclusively from the validation split, retaining only clips with valid video files. For each target step, the context history consists of up to the previous $L$ annotated step segments from the same video. State negatives are sampled from the same video but different step segments, with preference for temporally diverse mismatched steps, e.g., past, future, and additional non-target steps when available, up to 3 in total. When selecting past negatives, we avoid the immediate predecessor whenever possible, since it already appears in the context history. Identity negatives are mined globally across different videos using Sentence-BERT caption similarity. Specifically, we encode all step captions with \texttt{all-MiniLM-L6-v2}, compute cosine similarity between captions, and use the ranked cross-video neighbors as the identity-negative pool, from which up to 3 identity negatives are sampled per query. Easy negatives are sampled randomly from different videos.

\textbf{COIN.}
For COIN, we rely on the annotated procedural structure rather than external text mining. Context history is formed from preceding steps within the same video. State negatives are sampled from the same video but different steps. Identity negatives are sampled from different videos that share the same recipe/task category and step ID as the target step, yielding semantically matched but identity-inconsistent distractors. Easy negatives are sampled from clips with different recipe/task categories.

\textbf{CrossTask.}
For CrossTask, context history is again constructed from preceding steps within the same video. State negatives are sampled from the same video but different steps, with preference for past and future mismatched steps when available; as in YouCook2, we avoid using the immediate predecessor as a past negative whenever possible. Identity negatives are sampled from different videos within the same task and same step index as the target step. If insufficient such clips exist, we fall back to clips from different videos within the same task but different step indices. Easy negatives are sampled from different tasks.

Training-time candidate construction differs from the fixed evaluation benchmark above and is described separately below.

\paragraph{Training Candidate Construction and Hard Negatives.}
For training, each instance contains one ground-truth clip together with state and identity hard negatives. We sample up to 3 state negatives and up to 3 identity negatives per instance. If one negative pool is insufficient, we fall back to sampling from the other pool; if both are unavailable, we use zero vectors. This fallback strategy is used only for training-time instance construction. Evaluation instead follows the fixed 1-vs-9 benchmark protocol described in Sec.~\ref{subsec:eval_protocol} and Sec.~\ref{subsec:setup}, with up to 3 state negatives and up to 3 identity negatives per query, backfilled across hard-negative pools and then with easy negatives when needed.

\begin{table}[h]
\centering
\small
\caption{\textbf{Dataset-specific benchmark mining rules.}}
\label{tab:benchmark_mining}
\begin{tabular}{lccc}
\toprule
Dataset & State negatives & Identity negatives & Easy negatives \\
\midrule
YouCook2 & same video, diff step & SBERT top-3 cross-video clips & random diff video \\
COIN & same video, diff step & same recipe + same step, diff video & diff recipe \\
CrossTask & same video, diff step & same task + same step, diff video & diff task \\
\bottomrule
\end{tabular}
\end{table}

\paragraph{CAST Architecture Details.}
CAST is implemented as a lightweight residual adapter that operates in the frozen text-video embedding space.
Let $d$ denote the embedding dimension of the underlying frozen backbone (e.g., $d=512$ for CLIP-B/32).
Given the query text embedding $q_t \in \mathbb{R}^d$, the last context clip embedding $v_{t-1} \in \mathbb{R}^d$, and a context history
$H_t = \{h_1, \dots, h_L\} \in \mathbb{R}^{L \times d}$, we first apply L2 normalization to all input embeddings.

CAST models the transition as the sum of two complementary paths,
$\Delta = \Delta_{\mathrm{cond}} + \Delta_{\mathrm{ctx}}$.
The \textbf{instruction-conditioned state-transition path} (Path 1 in \S\ref{subsec:cast_model}) takes the concatenation $[q_t; v_{t-1}] \in \mathbb{R}^{2d}$ as input and applies a two-layer MLP:
$\mathrm{Linear}(2d, 2d) \rightarrow \mathrm{LayerNorm}(2d) \rightarrow \mathrm{ReLU} \rightarrow \mathrm{Dropout}(0.1) \rightarrow \mathrm{Linear}(2d, d)$.
This yields the transition component $\Delta_{\mathrm{cond}} \in \mathbb{R}^d$.

The context path first projects the query and context features through linear layers,
$q'_t = W_q q_t$ and $H'_t = W_h H_t$, where $W_q, W_h \in \mathbb{R}^{d \times d}$.
We then apply a single multi-head cross-attention layer with 8 heads, using the projected query as the attention query and the projected context sequence as keys and values.
The resulting attended feature is passed through a residual MLP:
$\mathrm{LayerNorm}(d) \rightarrow \mathrm{Linear}(d, d) \rightarrow \mathrm{ReLU} \rightarrow \mathrm{Linear}(d, d)$,
yielding $\Delta_{\mathrm{ctx}} \in \mathbb{R}^d$.
For variable-length context histories, we left-pad the sequence with zeros and apply a key-padding mask in attention.

The final predicted next-state embedding is computed via a direct residual update:
\[
\hat{v}_t = \mathrm{Norm}\!\left(v_{t-1} + \Delta_{\mathrm{cond}} + \Delta_{\mathrm{ctx}}\right),
\]
where $\mathrm{Norm}(\cdot)$ denotes L2 normalization.
Unless otherwise stated, we do not use an additional gating module in the main model.

\begin{table}[h]
\centering
\small
\caption{\textbf{CAST module specification.}}
\label{tab:cast_arch}
\begin{tabular}{l l}
\toprule
Item & Specification \\
\midrule
Embedding dimension $d$ & backbone-dependent (512 for CLIP-B/32) \\
Input normalization & L2 normalization \\
Conditioned path & Linear$(2d,2d)$ + LN + ReLU + Dropout(0.1) + Linear$(2d,d)$ \\
Context projections & Linear$(d,d)$ for query and context \\
Cross-attention & 1 layer, 8 heads \\
Context MLP & LN + Linear$(d,d)$ + ReLU + Linear$(d,d)$ \\
Residual update & $\hat v_t = \mathrm{Norm}(v_{t-1} + \Delta_{\mathrm{cond}} + \Delta_{\mathrm{ctx}})$ \\
Gating & none \\
\bottomrule
\end{tabular}
\end{table}

\paragraph{Optimization.}
We train CAST with AdamW (lr $=10^{-4}$, weight decay $=10^{-3}$) using batch size 512.
The number of training epochs is both dataset- and backbone-dependent.
For the main CLIP-based experiments, we train for 30, 20, and 50 epochs on YouCook2, COIN, and CrossTask, respectively.
For the alternative frozen backbones (InternVideo2, VideoPrism, GME-Qwen2-VL-2B, and Qwen3-VL-Embedding-2B),
we train for 30, 30, and 50 epochs on YouCook2, COIN, and CrossTask, respectively.
We use the Type-Aware contrastive objective defined in Sec.~\ref{subsec:training_inference}, with temperature $\tau=0.07$ and loss weights $\lambda_s=5.0$ and $\lambda_i=1.0$.

\paragraph{Inference Details.}
At inference time, we use the three score components and the Full Ensemble rule defined in Sec.~\ref{subsec:training_inference}.
In the main CLIP-based experiments, these scores are computed in the frozen CLIP embedding space.
For the universality results in Table~\ref{tab:universality}, the same score definitions are applied analogously within each backbone's native frozen text-video embedding space: $A$ denotes the native text-video similarity, $B$ denotes the similarity between the last context clip and the candidate clip, and $C$ denotes the similarity between the CAST-predicted next-state embedding and the candidate clip.
For the Full Ensemble score in Eq.~\eqref{eq:full_ensemble}, the ensemble coefficients are selected on a held-out validation split for each dataset/backbone setting using a fixed grid search with $w_v \in \{0.0, 0.1, \dots, 0.5\}$ and $w_p \in \{0.2, 0.3, \dots, 1.5\}$.
Once selected, these coefficients are frozen and applied to all reported results within that setting.
We do not tune coefficients per query, per example, or on the evaluation split.

\begin{table}[t]
\centering
\small
\caption{\textbf{CAST training and inference settings.}
Optimizer, batch size, temperature, and the overall inference formulation are shared across experiments, while selected training and inference details vary by dataset and backbone setting.}
\label{tab:appendix_cast_settings}
\vspace{-4pt}
\setlength{\tabcolsep}{5pt}
\begin{tabular}{l|p{8.5cm}}
\toprule
\textbf{Item} & \textbf{Setting} \\
\midrule
Context length $L$ & 5 (left-pad with zeros; masked attention) \\
Training epochs & CLIP: 30 / 20 / 50 on YouCook2 / COIN / CrossTask; InternVideo2 / VideoPrism / GME-Qwen2-VL-2B / Qwen3-VL-Embedding-2B: 30 / 30 / 50 \\
Optimizer & AdamW (lr $10^{-4}$, wd $10^{-3}$) \\
Batch size & 512 \\
Temperature $\tau$ & 0.07 \\
Hard negatives (training) & up to 3 state + 3 identity per instance \\
Evaluation pool & fixed 1 GT + 9 negatives; target composition is up to 3 state + up to 3 identity, with backfill across hard-negative pools and easy negatives used to fill remaining slots \\
Loss weights & $\lambda_s=5.0$ (state), $\lambda_i=1.0$ (identity) \\
Ensemble form & Full Ensemble: $S = A + w_v B + w_p C$ \\
\bottomrule
\end{tabular}
\vspace{-6pt}
\end{table}

\section{Appendix: Additional Qualitative Results and Evaluation Details}
\label{sec:qualitative_appendix}

\subsection{Human Evaluation Protocol for Generation}
\label{app:human_eval_generation}

To evaluate whether CAST improves candidate selection for black-box video generation, we conduct a blind human study on 300 prompts randomly sampled from the YouCook2 validation benchmark using a fixed random seed. Each prompt is constructed from a benchmark query by pairing the query instruction text with a single visual context image. Specifically, we extract the context image as the last frame of the most recent clip in the query's context history.

For generation, we use Veo to produce $K=4$ candidate continuation videos per prompt. The generation prompt follows a simple template:
\textit{``This image shows the context of the previous step. Generate a video for the next step: [instruction].''}
We intentionally keep this prompt simple and rely on CAST at reranking time, rather than introducing additional handcrafted prompt constraints.

\paragraph{Generation Setup and Reproducibility.}
We use the Veo API to generate candidate continuation videos from the same context image and instruction prompt. For each prompt, we obtain $K=4$ candidates from repeated generation calls using different random seeds $\{0,1,2,3\}$. Unless otherwise specified, all other generation parameters follow the API defaults. We use the \texttt{veo-3.1-fast-generate-preview} model with 720p resolution, 16:9 aspect ratio, and 8-second duration. If a generation call fails or returns an invalid output, we retry until a valid output is obtained. In one case, generation repeatedly failed for a prompt despite multiple retries, likely due to the API safety filter. That prompt was discarded and replaced by a newly sampled prompt, so that the final human evaluation was conducted on 300 valid prompts.

We then apply two ranking strategies to the same candidate pool:
(1) \textbf{Standard Text Match}, which ranks candidates using only the text-video matching score with the query instruction; and
(2) \textbf{CAST Reranking}, which applies the same final reranking strategy as in retrieval, combining semantic matching, visual continuity, and CAST-predicted next-state compatibility. For each method, we retain the top-ranked candidate as its final selected output.

When the two methods select different candidates, annotators are shown the two selected videos in randomized order without method identities. We use three independent annotators for each prompt. Annotators compare the two videos under the following criteria:

\begin{itemize}
    \item \textbf{Overall Preference}: which video is better overall as the continuation of the given context and instruction.
    \item \textbf{Physical Plausibility}: whether the object interactions, motion, and scene dynamics remain physically coherent.
    \item \textbf{Temporal Logic}: whether the generated continuation follows the procedural state implied by the visual context and instruction.
\end{itemize}

For each criterion, annotators select one of four options: \emph{Video A}, \emph{Video B}, \emph{Tie}, or \emph{Cannot judge}. We aggregate judgments using majority vote across the three annotators and report preference rates over all non-overlap prompts. If no majority exists (e.g., \emph{A} / \emph{B} / \emph{Tie}), we record the outcome as a human tie. In the final study, no prompt was marked as \emph{Cannot judge} after aggregation.

When both ranking methods select the same candidate from the $K=4$ generated videos, we record the case as an \textbf{overlap}. In our evaluation, this occurs for 142 out of 300 prompts (47.3\%), leaving 158 non-overlap prompts (52.7\%) for pairwise preference aggregation. Such cases are excluded from pairwise preference aggregation, since identical outputs admit no meaningful comparison. Equivalently, CAST changes the final selected candidate on 158/300 prompts (52.7\%), defining the maximal subset on which pairwise human comparison is meaningful. We report the main results on this non-overlap subset in Table~\ref{tab:generation}.

\subsection{Qualitative Analysis of Generation Reranking}

\begin{figure*}[t]
    \centering
    \includegraphics[width=\linewidth]{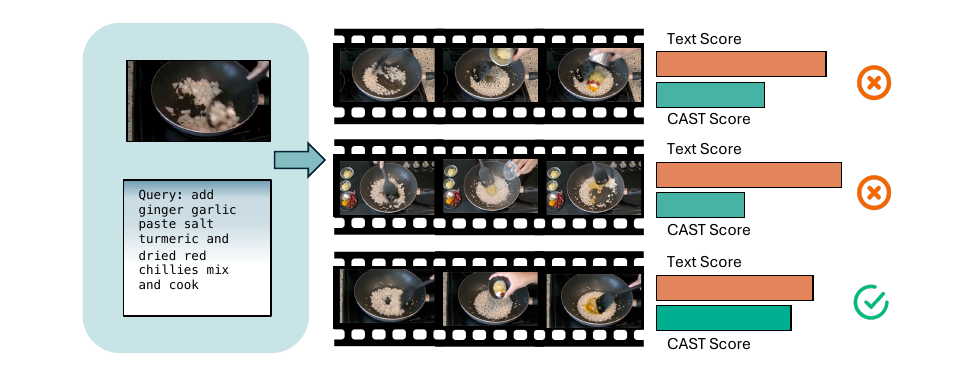}
    \caption{\textbf{Qualitative example of generation reranking.} Given a visual context and next-step instruction, text-only ranking is distracted by semantic similarity and consequently over-ranks procedurally mismatched candidates, whereas CAST more faithfully captures the expected next procedural state and selects a more coherent continuation.}
    \label{fig:qualitative_app}
\end{figure*}

\begin{figure*}[t]
    \centering
    \includegraphics[width=\linewidth]{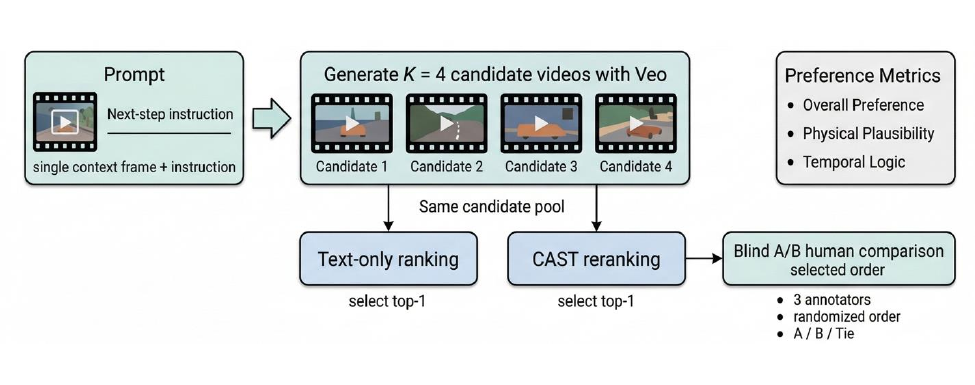}
    \caption{\textbf{Human evaluation protocol for generation reranking.} For each prompt, Veo generates $K=4$ candidate videos from a single context frame and a next-step instruction. Standard text matching and CAST reranking each select one top-ranked output from the same candidate pool, and annotators then compare the two selected videos in a blind A/B setting using three annotators, randomized order, and the options \emph{A}, \emph{B}, \emph{Tie}, or \emph{Cannot judge}.}
    \label{fig:human_eval_protocol}
\end{figure*}

We visualize a qualitative generation reranking example in Figure~\ref{fig:qualitative_app} and summarize the corresponding human evaluation protocol in Figure~\ref{fig:human_eval_protocol}. Consider the query \textit{``add ginger garlic paste''} following a visual context of frying onions. The baseline text score is often distracted by semantic overlap and consequently over-ranks candidates in which the ingredients already appear mixed or the procedural stage is mismatched, leading to \textbf{state errors}.

In contrast, \textbf{CAST} more reliably identifies the correct continuation by explicitly modeling the visual state transition $\Delta$. By extrapolating the current state $v_{t-1}$ under the query instruction, CAST captures the causal constraint induced by the action and favors candidates that match the expected next procedural state rather than scenes that merely contain semantically related objects.

Figure~\ref{fig:human_eval_protocol} illustrates the human evaluation setup used in Table~\ref{tab:generation}. For each prompt, Veo generates $K=4$ candidate videos from the same context frame and instruction. Standard text matching and CAST reranking then each select one top-ranked candidate from the same generated pool, and annotators compare the two selected outputs in a blind A/B setting.

%% file: main.bbl
\begin{thebibliography}{41}
\providecommand{\natexlab}[1]{#1}
\providecommand{\url}[1]{\texttt{#1}}
\expandafter\ifx\csname urlstyle\endcsname\relax
  \providecommand{\doi}[1]{doi: #1}\else
  \providecommand{\doi}{doi: \begingroup \urlstyle{rm}\Url}\fi

\bibitem[Alayrac et~al.(2016)Alayrac, Bojanowski, Agrawal, Sivic, Laptev, and Lacoste-Julien]{alayrac2016unsupervised}
J.-B. Alayrac, P.~Bojanowski, N.~Agrawal, J.~Sivic, I.~Laptev, and S.~Lacoste-Julien.
\newblock Unsupervised learning from narrated instruction videos.
\newblock In \emph{Proceedings of the IEEE conference on computer vision and pattern recognition}, pages 4575--4583, 2016.

\bibitem[Anne~Hendricks et~al.(2017)Anne~Hendricks, Wang, Shechtman, Sivic, Darrell, and Russell]{anne2017localizing}
L.~Anne~Hendricks, O.~Wang, E.~Shechtman, J.~Sivic, T.~Darrell, and B.~Russell.
\newblock Localizing moments in video with natural language.
\newblock In \emph{Proceedings of the IEEE international conference on computer vision}, pages 5803--5812, 2017.

\bibitem[Assran et~al.(2025)Assran, Bardes, Fan, Garrido, Howes, Muckley, Rizvi, Roberts, Sinha, Zholus, et~al.]{assran2025v}
M.~Assran, A.~Bardes, D.~Fan, Q.~Garrido, R.~Howes, M.~Muckley, A.~Rizvi, C.~Roberts, K.~Sinha, A.~Zholus, et~al.
\newblock V-jepa 2: Self-supervised video models enable understanding, prediction and planning.
\newblock \emph{arXiv preprint arXiv:2506.09985}, 2025.

\bibitem[Bain et~al.(2021)Bain, Nagrani, Varol, and Zisserman]{bain2021frozen}
M.~Bain, A.~Nagrani, G.~Varol, and A.~Zisserman.
\newblock Frozen in time: A joint video and image encoder for end-to-end retrieval.
\newblock In \emph{Proceedings of the IEEE/CVF international conference on computer vision}, pages 1728--1738, 2021.

\bibitem[Bardes et~al.(2024)Bardes, Garrido, Ponce, Chen, Rabbat, LeCun, Assran, and Ballas]{bardes2024revisiting}
A.~Bardes, Q.~Garrido, J.~Ponce, X.~Chen, M.~Rabbat, Y.~LeCun, M.~Assran, and N.~Ballas.
\newblock Revisiting feature prediction for learning visual representations from video.
\newblock \emph{arXiv preprint arXiv:2404.08471}, 2024.

\bibitem[Bengio et~al.(2013)Bengio, Courville, and Vincent]{bengio2013representation}
Y.~Bengio, A.~Courville, and P.~Vincent.
\newblock Representation learning: A review and new perspectives.
\newblock \emph{IEEE transactions on pattern analysis and machine intelligence}, 35\penalty0 (8):\penalty0 1798--1828, 2013.

\bibitem[Chang et~al.(2020)Chang, Huang, Xu, Adeli, Fei-Fei, and Niebles]{chang2020procedure}
C.-Y. Chang, D.-A. Huang, D.~Xu, E.~Adeli, L.~Fei-Fei, and J.~C. Niebles.
\newblock Procedure planning in instructional videos.
\newblock In \emph{European Conference on Computer Vision}, pages 334--350. Springer, 2020.

\bibitem[Chen and Dolan(2011)]{chen2011collecting}
D.~Chen and W.~B. Dolan.
\newblock Collecting highly parallel data for paraphrase evaluation.
\newblock In \emph{Proceedings of the 49th annual meeting of the association for computational linguistics: human language technologies}, pages 190--200, 2011.

\bibitem[Fu et~al.(2021)Fu, Li, Gan, Lin, Wang, Wang, and Liu]{fu2021violet}
T.-J. Fu, L.~Li, Z.~Gan, K.~Lin, W.~Y. Wang, L.~Wang, and Z.~Liu.
\newblock Violet: End-to-end video-language transformers with masked visual-token modeling.
\newblock \emph{arXiv preprint arXiv:2111.12681}, 2021.

\bibitem[Gao et~al.(2023)Gao, Han, Zhang, Lin, Geng, Zhou, Zhang, Lu, He, Yue, et~al.]{gao2023llama}
P.~Gao, J.~Han, R.~Zhang, Z.~Lin, S.~Geng, A.~Zhou, W.~Zhang, P.~Lu, C.~He, X.~Yue, et~al.
\newblock Llama-adapter v2: Parameter-efficient visual instruction model.
\newblock \emph{arXiv preprint arXiv:2304.15010}, 2023.

\bibitem[Geirhos et~al.(2020)Geirhos, Jacobsen, Michaelis, Zemel, Brendel, Bethge, and Wichmann]{geirhos2020shortcut}
R.~Geirhos, J.-H. Jacobsen, C.~Michaelis, R.~Zemel, W.~Brendel, M.~Bethge, and F.~A. Wichmann.
\newblock Shortcut learning in deep neural networks.
\newblock \emph{Nature Machine Intelligence}, 2\penalty0 (11):\penalty0 665--673, 2020.

\bibitem[{Google DeepMind}(2024)]{veo2024}
{Google DeepMind}.
\newblock Veo: Our most capable generative video model, 2024.
\newblock URL \url{https://deepmind.google/technologies/veo/}.
\newblock Accessed: 2024-05-14.

\bibitem[He et~al.(2016)He, Zhang, Ren, and Sun]{he2016deep}
K.~He, X.~Zhang, S.~Ren, and J.~Sun.
\newblock Deep residual learning for image recognition.
\newblock In \emph{Proceedings of the IEEE conference on computer vision and pattern recognition}, pages 770--778, 2016.

\bibitem[Houlsby et~al.(2019)Houlsby, Giurgiu, Jastrzebski, Morrone, De~Laroussilhe, Gesmundo, Attariyan, and Gelly]{houlsby2019parameter}
N.~Houlsby, A.~Giurgiu, S.~Jastrzebski, B.~Morrone, Q.~De~Laroussilhe, A.~Gesmundo, M.~Attariyan, and S.~Gelly.
\newblock Parameter-efficient transfer learning for nlp.
\newblock In \emph{International conference on machine learning}, pages 2790--2799. PMLR, 2019.

\bibitem[Hu et~al.(2022)Hu, Shen, Wallis, Allen-Zhu, Li, Wang, Wang, Chen, et~al.]{hu2022lora}
E.~J. Hu, Y.~Shen, P.~Wallis, Z.~Allen-Zhu, Y.~Li, S.~Wang, L.~Wang, W.~Chen, et~al.
\newblock Lora: Low-rank adaptation of large language models.
\newblock \emph{ICLR}, 1\penalty0 (2):\penalty0 3, 2022.

\bibitem[Isola et~al.(2015)Isola, Lim, and Adelson]{isola2015discovering}
P.~Isola, J.~J. Lim, and E.~H. Adelson.
\newblock Discovering states and transformations in image collections.
\newblock In \emph{Proceedings of the IEEE conference on computer vision and pattern recognition}, pages 1383--1391, 2015.

\bibitem[Li et~al.(2020)Li, Chen, Cheng, Gan, Yu, and Liu]{li2020hero}
L.~Li, Y.-C. Chen, Y.~Cheng, Z.~Gan, L.~Yu, and J.~Liu.
\newblock Hero: Hierarchical encoder for video+ language omni-representation pre-training.
\newblock \emph{arXiv preprint arXiv:2005.00200}, 2020.

\bibitem[Li et~al.(2026)Li, Zhang, Long, Keqin, Song, Bai, Yang, Xie, Yang, Liu, Zhou, and Lin]{qwen3vlembedding}
M.~Li, Y.~Zhang, D.~Long, C.~Keqin, S.~Song, S.~Bai, Z.~Yang, P.~Xie, A.~Yang, D.~Liu, J.~Zhou, and J.~Lin.
\newblock Qwen3-vl-embedding and qwen3-vl-reranker: A unified framework for state-of-the-art multimodal retrieval and ranking.
\newblock \emph{arXiv preprint arXiv:2601.04720}, 2026.

\bibitem[Loshchilov and Hutter(2017)]{loshchilov2017decoupled}
I.~Loshchilov and F.~Hutter.
\newblock Decoupled weight decay regularization.
\newblock \emph{arXiv preprint arXiv:1711.05101}, 2017.

\bibitem[Luo et~al.(2020)Luo, Ji, Shi, Huang, Duan, Li, Li, Bharti, and Zhou]{luo2020univl}
H.~Luo, L.~Ji, B.~Shi, H.~Huang, N.~Duan, T.~Li, J.~Li, T.~Bharti, and M.~Zhou.
\newblock Univl: A unified video and language pre-training model for multimodal understanding and generation.
\newblock \emph{arXiv preprint arXiv:2002.06353}, 2020.

\bibitem[Luo et~al.(2021)Luo, Ji, Zhong, Chen, Lei, Duan, and Li]{luo2021clip4clip}
H.~Luo, L.~Ji, M.~Zhong, Y.~Chen, W.~Lei, N.~Duan, and T.~Li.
\newblock Clip4clip: An empirical study of clip for end to end video clip retrieval.
\newblock \emph{arXiv preprint arXiv:2104.08860}, 2021.

\bibitem[Miech et~al.(2019)Miech, Zhukov, Alayrac, Tapaswi, Laptev, and Sivic]{miech2019howto100m}
A.~Miech, D.~Zhukov, J.-B. Alayrac, M.~Tapaswi, I.~Laptev, and J.~Sivic.
\newblock Howto100m: Learning a text-video embedding by watching hundred million narrated video clips.
\newblock In \emph{Proceedings of the IEEE/CVF international conference on computer vision}, pages 2630--2640, 2019.

\bibitem[OpenAI(2024)]{openai2024sora}
OpenAI.
\newblock Video generation models as world simulators.
\newblock \url{https://openai.com/research/video-generation-models-as-world-simulators}, 2024.
\newblock Accessed: 2024-02-15.

\bibitem[Radford et~al.(2021)Radford, Kim, Hallacy, Ramesh, Goh, Agarwal, Sastry, Askell, Mishkin, Clark, et~al.]{radford2021learning}
A.~Radford, J.~W. Kim, C.~Hallacy, A.~Ramesh, G.~Goh, S.~Agarwal, G.~Sastry, A.~Askell, P.~Mishkin, J.~Clark, et~al.
\newblock Learning transferable visual models from natural language supervision.
\newblock In \emph{International conference on machine learning}, pages 8748--8763. PMLR, 2021.

\bibitem[Reimers and Gurevych(2019)]{reimers2019sentence}
N.~Reimers and I.~Gurevych.
\newblock Sentence-bert: Sentence embeddings using siamese bert-networks.
\newblock \emph{arXiv preprint arXiv:1908.10084}, 2019.

\bibitem[Sou{\v{c}}ek et~al.(2022)Sou{\v{c}}ek, Alayrac, Miech, Laptev, and Sivic]{souvcek2022look}
T.~Sou{\v{c}}ek, J.-B. Alayrac, A.~Miech, I.~Laptev, and J.~Sivic.
\newblock Look for the change: Learning object states and state-modifying actions from untrimmed web videos.
\newblock In \emph{Proceedings of the IEEE/CVF Conference on Computer Vision and Pattern Recognition}, pages 13956--13966, 2022.

\bibitem[Sun et~al.(2019)Sun, Myers, Vondrick, Murphy, and Schmid]{sun2019videobert}
C.~Sun, A.~Myers, C.~Vondrick, K.~Murphy, and C.~Schmid.
\newblock Videobert: A joint model for video and language representation learning.
\newblock In \emph{Proceedings of the IEEE/CVF international conference on computer vision}, pages 7464--7473, 2019.

\bibitem[Tang et~al.(2019)Tang, Ding, Rao, Zheng, Zhang, Zhao, Lu, and Zhou]{tang2019coin}
Y.~Tang, D.~Ding, Y.~Rao, Y.~Zheng, D.~Zhang, L.~Zhao, J.~Lu, and J.~Zhou.
\newblock Coin: A large-scale dataset for comprehensive instructional video analysis.
\newblock In \emph{Proceedings of the IEEE/CVF Conference on Computer Vision and Pattern Recognition}, pages 1207--1216, 2019.

\bibitem[Team et~al.(2025)Team, Chen, Ci, Du, Feng, Gai, Guo, Han, He, He, et~al.]{team2025kling}
K.~Team, J.~Chen, Y.~Ci, X.~Du, Z.~Feng, K.~Gai, S.~Guo, F.~Han, J.~He, K.~He, et~al.
\newblock Kling-omni technical report.
\newblock \emph{arXiv preprint arXiv:2512.16776}, 2025.

\bibitem[Vaswani et~al.(2017)Vaswani, Shazeer, Parmar, Uszkoreit, Jones, Gomez, Kaiser, and Polosukhin]{vaswani2017attention}
A.~Vaswani, N.~Shazeer, N.~Parmar, J.~Uszkoreit, L.~Jones, A.~N. Gomez, {\L}.~Kaiser, and I.~Polosukhin.
\newblock Attention is all you need.
\newblock \emph{Advances in neural information processing systems}, 30, 2017.

\bibitem[Villegas et~al.(2017)Villegas, Yang, Hong, Lin, and Lee]{villegas2017decomposing}
R.~Villegas, J.~Yang, S.~Hong, X.~Lin, and H.~Lee.
\newblock Decomposing motion and content for natural video sequence prediction.
\newblock \emph{arXiv preprint arXiv:1706.08033}, 2017.

\bibitem[Villegas et~al.(2022)Villegas, Babaeizadeh, Kindermans, Moraldo, Zhang, Saffar, Castro, Kunze, and Erhan]{villegas2022phenaki}
R.~Villegas, M.~Babaeizadeh, P.-J. Kindermans, H.~Moraldo, H.~Zhang, M.~T. Saffar, S.~Castro, J.~Kunze, and D.~Erhan.
\newblock Phenaki: Variable length video generation from open domain textual description.
\newblock \emph{arXiv preprint arXiv:2210.02399}, 2022.

\bibitem[Wang et~al.(2018)Wang, Jiang, Ma, Liu, and Xu]{wang2018bidirectional}
J.~Wang, W.~Jiang, L.~Ma, W.~Liu, and Y.~Xu.
\newblock Bidirectional attentive fusion with context gating for dense video captioning.
\newblock In \emph{Proceedings of the IEEE conference on computer vision and pattern recognition}, pages 7190--7198, 2018.

\bibitem[Wang et~al.(2022)Wang, Li, Li, He, Huang, Zhao, Zhang, Xu, Liu, Wang, et~al.]{wang2022internvideo}
Y.~Wang, K.~Li, Y.~Li, Y.~He, B.~Huang, Z.~Zhao, H.~Zhang, J.~Xu, Y.~Liu, Z.~Wang, et~al.
\newblock Internvideo: General video foundation models via generative and discriminative learning.
\newblock \emph{arXiv preprint arXiv:2212.03191}, 2022.

\bibitem[Wang et~al.(2024)Wang, Li, Li, Yu, He, Chen, Pei, Zheng, Wang, Shi, et~al.]{wang2024internvideo2}
Y.~Wang, K.~Li, X.~Li, J.~Yu, Y.~He, G.~Chen, B.~Pei, R.~Zheng, Z.~Wang, Y.~Shi, et~al.
\newblock Internvideo2: Scaling foundation models for multimodal video understanding.
\newblock In \emph{European Conference on Computer Vision}, pages 396--416. Springer, 2024.

\bibitem[Xu et~al.(2016)Xu, Mei, Yao, and Rui]{xu2016msr}
J.~Xu, T.~Mei, T.~Yao, and Y.~Rui.
\newblock Msr-vtt: A large video description dataset for bridging video and language.
\newblock In \emph{Proceedings of the IEEE conference on computer vision and pattern recognition}, pages 5288--5296, 2016.

\bibitem[Yan et~al.(2022)Yan, Zhu, Wang, Cao, Zhang, Ghosh, Wu, and Yu]{yan2022videococa}
S.~Yan, T.~Zhu, Z.~Wang, Y.~Cao, M.~Zhang, S.~Ghosh, Y.~Wu, and J.~Yu.
\newblock Videococa: Video-text modeling with zero-shot transfer from contrastive captioners.
\newblock \emph{arXiv preprint arXiv:2212.04979}, 2022.

\bibitem[Zhang et~al.(2024)Zhang, Zhang, Xie, Li, Dai, Long, Xie, Zhang, Li, and Zhang]{zhang2024gme}
X.~Zhang, Y.~Zhang, W.~Xie, M.~Li, Z.~Dai, D.~Long, P.~Xie, M.~Zhang, W.~Li, and M.~Zhang.
\newblock Gme: Improving universal multimodal retrieval by multimodal llms.
\newblock \emph{arXiv preprint arXiv:2412.16855}, 2024.

\bibitem[Zhao et~al.(2024)Zhao, Gundavarapu, Yuan, Zhou, Yan, Sun, Friedman, Qian, Weyand, Zhao, et~al.]{zhao2024videoprism}
L.~Zhao, N.~B. Gundavarapu, L.~Yuan, H.~Zhou, S.~Yan, J.~J. Sun, L.~Friedman, R.~Qian, T.~Weyand, Y.~Zhao, et~al.
\newblock Videoprism: A foundational visual encoder for video understanding.
\newblock \emph{arXiv preprint arXiv:2402.13217}, 2024.

\bibitem[Zhou et~al.(2018)Zhou, Xu, and Corso]{zhou2018towards}
L.~Zhou, C.~Xu, and J.~Corso.
\newblock Towards automatic learning of procedures from web instructional videos.
\newblock In \emph{Proceedings of the AAAI conference on artificial intelligence}, volume~32, 2018.

\bibitem[Zhukov et~al.(2019)Zhukov, Alayrac, Cinbis, Fouhey, Laptev, and Sivic]{zhukov2019cross}
D.~Zhukov, J.-B. Alayrac, R.~G. Cinbis, D.~Fouhey, I.~Laptev, and J.~Sivic.
\newblock Cross-task weakly supervised learning from instructional videos.
\newblock In \emph{Proceedings of the IEEE/CVF Conference on Computer Vision and Pattern Recognition}, pages 3537--3545, 2019.

\end{thebibliography}
